\documentclass{article}

\newif\ificlr
\iclrfalse

\ificlr
  \usepackage{iclr2027_conference,times}
\else
  \usepackage{arxiv}
\fi

\usepackage{microtype}
\usepackage{hyperref}
\usepackage{url}
\usepackage{xurl}
\usepackage{amsmath}
\usepackage{amsfonts}
\usepackage{booktabs}
\usepackage[capitalise]{cleveref} %
\crefformat{section}{\S#2#1#3}
\Crefformat{section}{\S#2#1#3}
\crefformat{subsection}{\S#2#1#3}
\Crefformat{subsection}{\S#2#1#3}
\crefmultiformat{section}{\S#2#1#3}{ and \S#2#1#3}{, \S#2#1#3}{ and \S#2#1#3}
\Crefmultiformat{section}{\S#2#1#3}{ and \S#2#1#3}{, \S#2#1#3}{ and \S#2#1#3}
\usepackage{graphicx}
\graphicspath{{./}{../}}
\usepackage{amssymb}
\usepackage{algorithm}
\usepackage[noend]{algorithmic}
\usepackage{wrapfig}
\usepackage[normalem]{ulem}
\usepackage{pifont}
\usepackage{tcolorbox, listings, bm}
\usepackage{enumitem}
\tcbuselibrary{skins, breakable, listings, raster}

\usepackage{lineno}
\usepackage{xspace}
\definecolor{scggreen}{HTML}{DCEEF0}
\definecolor{darkblue}{rgb}{0, 0, 0.5}

\hypersetup{colorlinks=true, citecolor=darkblue, linkcolor=darkblue, urlcolor=darkblue}

\title{Hinted Scaling Laws}

\author{Suze van Adrichem*, Luke Bailey*, Rohith Kuditipudi*, Christine Ye, Neil Band\\[4pt]
Ahmed Ahmed, John Duchi, Percy Liang, Tatsunori Hashimoto \\[12pt]
\textnormal{Stanford University}}

\usepackage{xcolor}
\definecolor{infra-blue}{HTML}{0076BA}
\definecolor{infra-fault}{HTML}{F27300}
\newcommand{\LB}[1]{\textcolor{red}{[LB: #1]}}
\newcommand{\TBH}[1]{\textcolor{blue}{[TH: #1]}}
\newcommand{\SV}[1]{\textcolor{cyan}{[SV: #1]}}
\newcommand{\NB}[1]{\textcolor{purple}{[NB: #1]}}
\newcommand{\RK}[1]{\textcolor{orange}{[RK: #1]}}
\newcommand{\pl}[1]{\textcolor{purple}{[PL: #1]}}

\newif\ifhidecomments
\hidecommentsfalse   %

\ifhidecomments
  \makeatletter
  \renewcommand{\LB}[1]{\@bsphack\@esphack}
  \renewcommand{\TBH}[1]{\@bsphack\@esphack}
  \renewcommand{\SV}[1]{\@bsphack\@esphack}
  \renewcommand{\NB}[1]{\@bsphack\@esphack}
  \renewcommand{\RK}[1]{\@bsphack\@esphack}
  \renewcommand{\pl}[1]{\@bsphack\@esphack}
  \makeatother
\fi

\newcommand{\method}{benchmark hinting}
\newcommand{\Method}{Benchmark hinting}
\usepackage{amssymb}   %
\usepackage{xcolor}

\definecolor{wintick}{HTML}{63A95E}    %
\definecolor{meancol}{HTML}{D1495B}    %
\definecolor{methcol}{HTML}{A9A8A4}    %
\definecolor{basecol}{HTML}{0B0B0B}    %
\definecolor{traincol}{HTML}{52514E}   %
\definecolor{testcol}{HTML}{EDA100}    %
\definecolor{fcstcol}{HTML}{C0392B}    %
\definecolor{refcol}{HTML}{8AA8C4}     %
\definecolor{pkcol}{HTML}{3B7DD8}      %

\definecolor{avgkcol}{HTML}{000000}     %
\definecolor{hintshrinkcol}{HTML}{9467BD}    %
\definecolor{loglosscol}{HTML}{B22222}  %
\definecolor{chancecol}{HTML}{808080}   %
\definecolor{hintlinecol}{HTML}{25848E} %
\DeclareRobustCommand{\hintgradient}{\textbf{\textcolor{hintlinecol}{semi-transparent lines}}}

\newcommand{\wintick}{\textcolor{wintick}{\checkmark}}

\newcommand{\traindot}{\textcolor{traincol}{$\bullet$}}
\newcommand{\testdot}{\textcolor{testcol}{$\bullet$}}

\usepackage{wrapfig}
\usepackage{booktabs}

\usepackage{amsmath,amsfonts,bm}

\def\eqref#1{equation~\ref{#1}}

\def\1{\bm{1}}

\DeclareMathAlphabet{\mathsfit}{\encodingdefault}{\sfdefault}{m}{sl}
\SetMathAlphabet{\mathsfit}{bold}{\encodingdefault}{\sfdefault}{bx}{n}

\begin{document}

\maketitle
{\renewcommand{\thefootnote}{}%
\footnotetext{%
\hspace{-1.8em}Correspondence to 
\href{mailto:suzeva@cs.stanford.edu,ljbailey@cs.stanford.edu,rohithk@cs.stanford.edu}{\texttt{\{suzeva,ljbailey,rohithk\}@cs.stanford.edu}} \\
\hspace{2em}Code and data available at  \url{https://github.com/tatsu-lab/hinted-scaling-laws}\\
* Equal contribution, alphabetical ordering
}}

\begin{abstract}
Scaling laws predict the outcome of large, expensive training runs by extrapolating trends from smaller, cheaper runs.
On downstream accuracy evaluations,
these predictions fail when the weaker models produced by the cheaper runs score near zero accuracy,
because these scores give no trend to extrapolate to larger scales.
We introduce hinted scaling laws, which use privileged information to 
turn a benchmark into a continuum of easier versions so that
even the small 
models
score above zero.
We find that accuracy rises smoothly and predictably with the hint
level, enabling new types of scaling laws that operate below the noise threshold.
On the Delphi pretraining ladder, a hinted scaling law fit on models trained with at most $10^{20}$ FLOPs forecasts the GSM8K accuracy of a $10^{23}$ FLOP model (25.9\%) to within 3 percentage points; a standard scaling law under-forecasts it by 13 percentage points when fit on the same models and by 12 even with models up to $10^{22}$ FLOPs in the fit.
Across both this pretraining ladder and a collection 
of post-trained model families, hinted scaling laws reduce the median forecasting error on GSM8K, AIME and HLE from 21 percentage points to 4.3. 
Hinting also reduces the number of test examples needed to tell whether one model is better than another by widening the
accuracy gaps between models while approximately preserving their order:
a single hinted rollout per problem ranks nearly indistinguishable models
as accurately as five unhinted rollouts.
Our results suggest that hinting is a general primitive for improving the 
accuracy and cost of downstream scaling laws and other prediction tasks
in model development.
\end{abstract}

\section{Introduction}

By extrapolating the effect of a design choice from small, cheap training runs to an expensive target scale, scaling laws \citep{kaplan2020scalinglawsneurallanguage, hoffmann2022trainingcomputeoptimallargelanguage} enable affordable tuning of data and algorithmic recipes to optimize metrics such as pretraining loss that vary predictably with scale.

Yet pretraining loss is an imperfect proxy for the downstream capabilities we care about, because lower loss does not reliably mean better task performance~\citep{liu2023same, springer2025overtrained, watts2026sharpness}.
Thus, we want scaling laws that directly predict downstream accuracy.
Such laws would let developers optimize design decisions for
downstream tasks directly
and help
safety teams anticipate the scale at which larger models acquire
dangerous capabilities.
Unfortunately, benchmark accuracy
does not vary predictably with scale like pretraining loss does.
Small models score near zero on challenging benchmarks, providing no signal for extrapolation~\citep{wei2022emergentabilitieslargelanguage}.
As a result, scaling laws fit directly to accuracy are often unreliable \citep{lourie2025scalinglawsunreliabledownstream}. 

\begin{figure}[t]
    \centering
    \includegraphics[width=\linewidth]{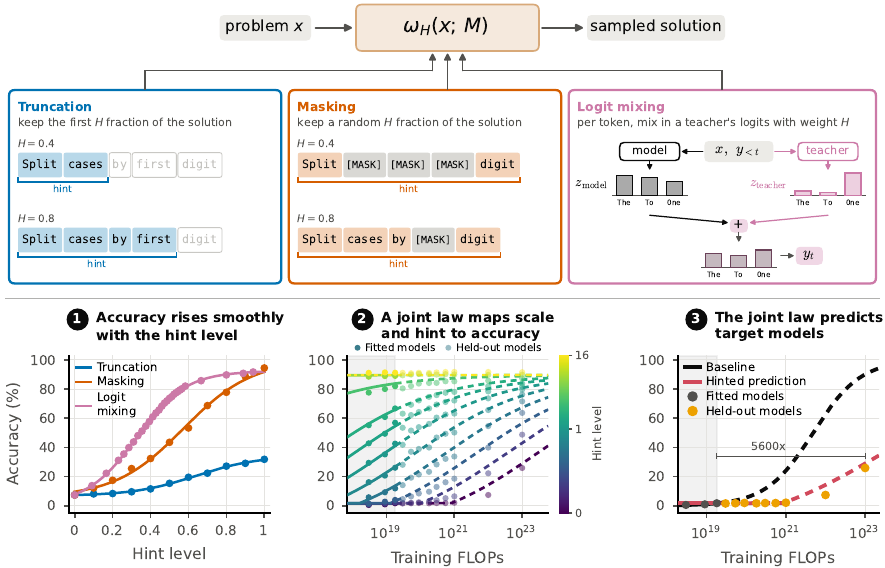}
    \vspace{-6mm}
    \caption{\textbf{Hinted scaling laws.}
          \textbf{Top:} \Method{} turns a benchmark
          into a continuous sequence of easier
          versions indexed by a hint level $H$,
          converting an answer sampling protocol $\omega$ into a hinted protocol
          $\omega_H$ by truncation, masking, or logit mixing.
          \textbf{Bottom:} results for the Delphi pretraining ladder on GSM8K.
          \textbf{(1)} Accuracy of the $10^{22}$ FLOPs model rises smoothly with $H$, where $H=0$ is the original benchmark.
          \textbf{(2)} A single joint law in model scale and hint level (\cref{eq:joint_law}) is fit to the three smallest models at every $H$ of logit mixing and extrapolates to the held-out larger models.
          \textbf{(3)} The same law, whose \textbf{\textcolor{traincol}{fitted models}} have 5600x less compute than the largest and score near zero at $H = 0$, is evaluated at $H = 0$ to give a \textbf{\textcolor{meancol}{hinted prediction}} that forecasts the \textbf{\textcolor{testcol}{held-out larger models}}. A \textbf{\textcolor{basecol}{baseline}} scaling law fit to the same models' unhinted accuracy alone overshoots.
        }
    \label{fig:overview}
\end{figure}

We find that benchmark accuracy does vary predictably with a second variable: how much \emph{help} the model is given. By modeling how accuracy varies with
both external help and model scale, we 
are able to produce reliable benchmark accuracy 
scaling laws, even when small models score near zero.
We formalize this help as \emph{\method{}}. 
Given a benchmark, \method{} constructs a sequence of easier benchmarks indexed by a hint level $H$, with the original benchmark at $H = 0$. 
We instantiate three hinting methods, which each make the benchmark 
easier using privileged information such as a worked solution to each problem (\cref{fig:overview}, top).

\Method{} produces two useful phenomena.
First, accuracy rises smoothly with $H$ even for models that score 
near zero accuracy on the original benchmark. %
Second, accuracy has a predictable joint structure over \emph{both} hint level $H$ and model scale, with increasing $H$ acting similarly to increasing model scale (\cref{sec:phenomena}).

Given this shared structure between model scale and hint level, we can apply \method{} to fit an accurate downstream accuracy scaling law even when the scaling ladder models score zero on the original benchmark (\cref{fig:overview}, bottom).
A standard scaling law fits benchmark accuracy as a function of model scale $C$ alone. 
We instead fit a joint law $L(C, H)$ over scale and hint level. 
This approach of bootstrapping a sequence of easier benchmarks ($H > 0$) from a challenging target benchmark ($H = 0$) enables the joint law $L(C, H)$ to extract useful information from smaller model scales that achieve zero performance on the original benchmark. 
To predict the accuracy on the original benchmark of 
a larger model with scale $C_\mathrm{large}$,
we simply evaluate $L(C_\mathrm{large}, 0)$.
Altogether, our contributions are as follows:

\begin{itemize}
  \interlinepenalty=10000 %
  \item \textbf{Methodology.} We introduce \method{}, which turns a benchmark into a 
     continuous sequence of progressively easier versions indexed by 
     a hint level, and instantiate it
     with three hinting methods: truncation, masking, and logit mixing (\cref{sec:preliminaries}).
   \item \textbf{Phenomena.} 
    We show that under each hinting method,  
   benchmark accuracy varies smoothly and predictably with the hint level, including for models that score near zero on the original benchmark.  We also find that 
     hint level $H$ and model scale form a simple joint structure, with 
     increasing hint level $H$ acting similarly to increasing model scale (\cref{sec:phenomena}).
   \item \textbf{Applications.} We use this structure in two ways.
     First, we fit a joint law over model scale and hint level to only
     the smallest models of pretrained and post-trained families. 
     Forecasting the
     largest model's accuracy on the original benchmark, over all model families and benchmarks we evaluate, it reaches a median absolute error of 4.3 percentage points
     against 21 for a standard scaling law. 
     For the Delphi pretraining ladder, the most granular model family we consider,
     a hinted scaling law that uses logit mixing with a shared teacher and is fit on models
     trained with at most $10^{20}$ FLOPs predicts the accuracy of a $10^{23}$ FLOP model
     to within 3 percentage points (\cref{sec:benchmark_performance}).
     A standard scaling law under-forecasts it by 13 percentage points, and by 12 even when
     fit on models of up to $10^{22}$ FLOPs.
     Second, because hinting widens the accuracy gaps between models while largely
     preserving their order, it ranks nearly indistinguishable models more accurately
     at small sampling budgets. With one truncation-hinted rollout per problem, our ranking
     improves Kendall's $\tau$ over an unhinted ranking by $0.33$ on AIME and $0.27$ on HLE
     (\cref{sec:model_ranking}).
\end{itemize}

Together, our results demonstrate that the primitive of 
\method{} is useful in a variety of prediction tasks, such as estimating benchmark accuracy at larger compute scale or with fewer samples. We anticipate that hinting will be more generally applicable, in settings such as predicting emergent capabilities on long-horizon agentic tasks or scaling down environments for predictable reinforcement learning scaling.

\section{Benchmark Hinting}
\label{sec:preliminaries}

We define a benchmark $B = (\mathcal{Q}, \omega)$ as a set of problems $\mathcal{Q}$, each with a way of
grading a solution, and a protocol $\omega$ that takes a model $M$ and a problem $x \in \mathcal{Q}$
and samples a solution. $\omega$ fixes the prompt format and sampling method.
We write $\mathrm{Acc}(M, B)$ for model $M$'s expected accuracy on $B$ over the randomness in $\omega$.
A \emph{\method{} method} maps $B$ and a \emph{hint level} $H \in [0, \infty)$ to a new
benchmark $B_H = (\mathcal{Q}, \omega_H)$ with the same problems but a protocol that makes
problems easier for the model as $H$ grows, for example by giving the model privileged information about the solution, giving the model more attempts, or adapting the model to the task.
In this work we instantiate \method{} 
by revealing part of a correct solution in the prompt or 
by letting a second stronger model influence token sampling.
We abbreviate $\mathrm{Acc}(M, H) := \mathrm{Acc}(M, B_H)$.

We outline three basic desiderata for a hinting method.
First, the benchmark at hint level $H = 0$ must match the original benchmark, so $\omega_0 = \omega$ and $B_0 = B$.
Second, $\mathrm{Acc}(M, H)$ should predictably increase in $H$ according to some simple law. 
Third, for a fixed hint level $H$, $\mathrm{Acc}(M, H)$ should increase predictably with the capability of the model $M$, measured by a scale variable $C$ such as $\log_{10}$ pretraining FLOPs or $\log_{10}$ parameter count. 
We refer to the latter two desiderata as \emph{hint scaling} and \emph{capability scaling}. 
In \cref{sec:methods}, we present instantiations of \method{}. 
In \cref{sec:phenomena}, we provide evidence that these methods 
demonstrate hint and capability scaling.
Then in \cref{sec:benchmark_performance}, we show 
that these methods can be used to 
construct a downstream scaling law 
that enables us to predict large-scale model performance.

\subsection{Instantiations of Benchmark Hinting}
\label{sec:methods}
We instantiate three \method{} methods. 
In each case, the method makes a problem easier 
by \emph{revealing privileged information to the model},
with $H$ controlling the amount of information revealed.
This privileged information takes 
the form of a reference correct solution to the problem, or information from a stronger
teacher model.
Thus it is
often a requirement of these methods 
that we have 
access to reference solutions.

\textbf{Truncation.}~~~~The protocol $\omega_H$ places the
first $H$ fraction of the reference solution, cut at a word boundary, in
the model's context before it answers. We reword the reference 
solution to not reveal the specific answer string (e.g. the numerical answer 
to an AIME problem).

\textbf{Masking.}~~~~We place
the reference solution in-context but replaces a randomly chosen
$1 - H$ fraction of its words with \texttt{[MASK]} so that at $H = 1$ the
solution is intact and at $H = 0$ every word is hidden. At $H=0$ we drop the hint entirely since it would consist only of \texttt{[MASK]} tokens.

\textbf{Logit mixing.}~~~~We decode the model jointly with a \emph{stronger teacher}, drawing the next 
token from 
some combination of the 
two models' logits. Concretely, the two models share a
vocabulary, and at each decoding step we 
write $z_{\mathrm{model}}(x, y_{<t})$ and
$z_{\mathrm{teacher}}(x, y_{<t})$
for the logits that the evaluated model and the teacher,
respectively, produce for the $t$-th answer position.
We use the following equation to mix the logits
\begin{equation}
\label{eq:logit_sum}
s_H(x, y_{<t})_j = z_{\mathrm{model}}(x, y_{<t})_j + H\, z_{\mathrm{teacher}}(x, y_{<t})_j,
\qquad H \in [0, \infty),
\end{equation}
and samples $y_t \sim \mathrm{softmax}(s_H / T)$.
 The sampled token is appended to both models' contexts and decoding
 continues. At $H = 0$ this
is the evaluated model alone. As $H$ increases, the teacher's logits
dominate. As $H \to \infty$ the sampling converges to greedy
sampling of the teacher, and accuracy is the same for every evaluated model.
The teacher can be any model that 
is meaningfully stronger on the benchmark 
than the student, whether this be 
because it is generally larger 
and more capable or because it is prompted 
with privileged information such 
as a reference solution. 

We can separate the \method{} methods 
into two categories based on
the requirements for running them.
Truncation and masking require only 
reference solutions to the benchmark.
Our main focus throughout the paper will be applying these two methods to a variety of families and benchmarks.
Logit mixing requires a good
teacher model. 
We
motivate its use in \cref{sec:logit-mixing} 
as a means of obtaining better forecasts from weaker models,
which we do using a pretraining ladder and prompting Llama-3.1-8B-Instruct with the solution in its context as the teacher.

\section{Hinting and Capability Scaling}
\label{sec:phenomena}

\begin{figure}[t]
    \centering
    \includegraphics[width=\linewidth]{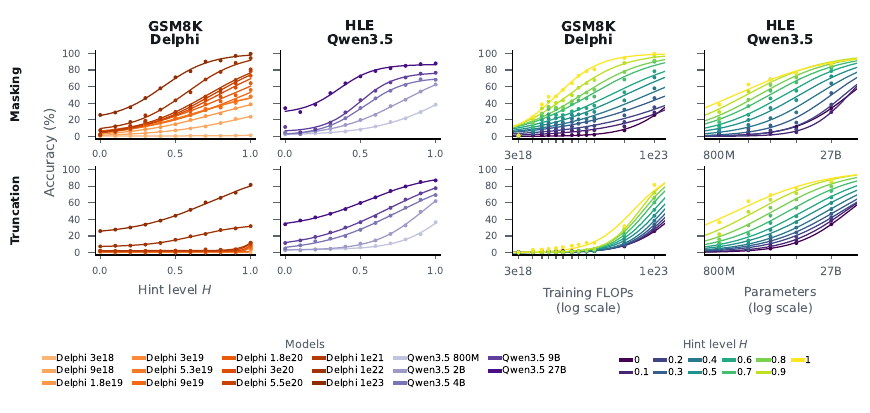}
    \vspace{-8mm}
    \caption{\textbf{Hinted benchmark accuracy varies smoothly in both
    hint level and scale.} Rows are masking and truncation. Columns
    are the Delphi pretraining ladder \citep{held2026delphi} on GSM8K and
    the Qwen3.5 family on HLE.
    \textbf{Left:} accuracy against $H$, one curve
    per model, darker is larger. \textbf{Right:} accuracy against model
    scale $C$, one curve per hint level, colored by $H$.  
    Each curve is a sigmoid fit to a single model scale or hint level. As a function of $H$ the asymptotes are free: $L + (A - L)\,\sigma(a_1 H + a_0)$, with
    $0 \leq L \leq A \leq 1$. As a function of $C$, they are pinned to 0 and 1:
    $\sigma(a_1 C + a_0)$, 
   since accuracy should approach 0 as scale shrinks and 1 as it grows.}
    \label{fig:phenomena_ladders}
\end{figure}

We study two core phenomena of \method{}.
First, accuracy increases 
smoothly as 
a function of the hint level $H$. 
Second, 
benchmark accuracy has a simple joint structure over 
both hint level and model scale
where increasing hint level acts
 similarly to increasing model scale.
We measure how benchmark accuracy varies with hint level
and model scale under masking and truncation (\cref{sec:methods}). Logit
mixing, which needs a stronger teacher, we study separately in
\cref{sec:logit-mixing}. To study the effect of hinting, 
we need model families, collections of
models trained in the same way that differ only in a scale variable $C$. 
We consider pretraining 
scaling ladders (where $C$ is $\log_{10}$ FLOPs) and released 
post-trained families (where $C$ is $\log_{10}$ parameters).

\begin{figure}[t]
    \centering
    \includegraphics[width=\linewidth]{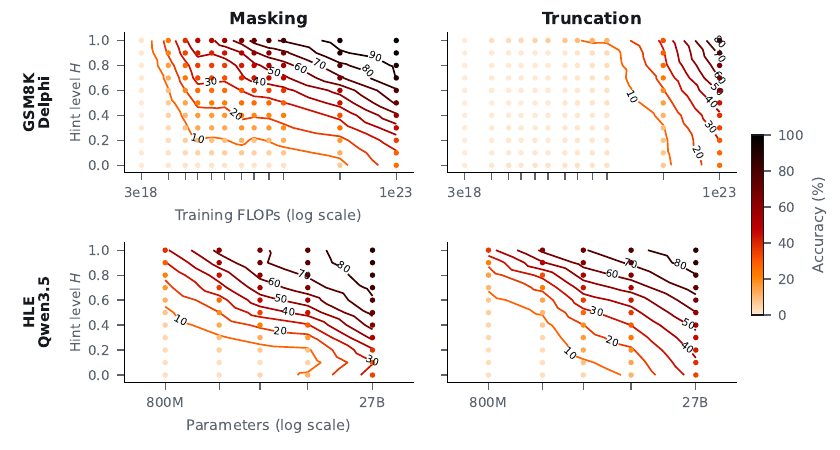}
    \vspace{-8mm}
    \caption{\textbf{Iso-accuracy contours in the scale and hint level
    plane.} Rows are the Delphi ladder on GSM8K and the Qwen3.5 family on
    HLE, columns masking and truncation. Each dot is one measured
    cell, colored by accuracy. Lines are contours of constant accuracy at
    10\% to 90\%, interpolated linearly between the cells. The slope of a contour is the exchange rate between scale and
    hint level at that accuracy. 
}
    \label{fig:iso_accuracy}
\end{figure}

\textbf{Methods and data.}~~~~For the pretraining ladder we use Delphi, 
an open-weight scaling ladder of
pretrained language models \citep{held2026delphi}. We use twelve
compute budgets, from $3 \times 10^{18}$ to $10^{23}$ FLOPs, spanning 447M
to 25B parameters. Every budget below $10^{23}$ FLOPs has three models
trained with different seeds, and the $10^{23}$ FLOPs budget has one. For
each Delphi budget we report accuracy averaged over its seeds, and we fit
all Delphi laws to these seed means. 
For post-trained families, we 
consider the
six dense Qwen2.5 Instruct models, six dense Qwen3 models, 
five dense Qwen3.5 models, and five Gemma-3 instruction-tuned
models. 
For brevity, we display figures in this section for the pretrained Delphi ladder and the
newest post-trained family, Qwen3.5 (see \cref{app:all_families} for the rest). 
We evaluate pretrained models on GSM8K and post-trained models on AIME 2025 and 2026
(AIME) and a subset of Humanity's Last Exam (HLE), described in \cref{app:benchmark_details}.

Truncation and masking require reference solutions. 
 For AIME and HLE
we generate these solutions as described in \cref{app:reference_solutions}. For GSM8K we use
the step-by-step solutions that come with the benchmark.
We sweep $H \in \{0, 0.1, \ldots, 1\}$.
At each $H$ we
draw 10 rollouts per problem on AIME and HLE and, on GSM8K, 32 per seed at every $H > 0$ and 64 per seed at $H = 0$,
 to estimate the expected accuracy over rollout stochasticity.

\textbf{Accuracy scales smoothly with hint level.}~~~~In \cref{fig:phenomena_ladders} (left) we show accuracies of 
Delphi on GSM8K and Qwen3.5 on HLE at various hint levels. 
For every model,
accuracy rises remarkably smoothly in $H$, and a sigmoid with
free asymptotes fits each model
very well. Larger models generally stay above smaller ones at every hint level, so \emph{the family's ordering is preserved as $H$ increases}.
The shape
of the rise is a property of the \method{} method. Masking climbs
steadily across the whole range, and truncation moves little until most of
the solution is shown.

\textbf{Accuracy of hinted benchmarks scales smoothly with compute.}~~~~The right half of \cref{fig:phenomena_ladders} fixes the hint level,
and we see at every level (1) the data follows a sigmoid closely, 
and (2) a higher hint level shifts the curve left 
with curves rarely crossing. Increasing $H$ thus moves rising accuracy to
  smaller models. For example, at $H=0$ 
the Delphi ladder sits at the noise floor until $10^{22}$ FLOPs, while under masking at $H = 0.5$
every model from $9 \times 10^{18}$ FLOPs upward scores above 7\%.

\textbf{Hint level and capability exchange rate.}~~~~
We now move on to analyzing how accuracy 
varies as compute and hint level change \emph{jointly}.
The two sets of slices in \cref{fig:phenomena_ladders} are cut from one surface, 
namely accuracy as a function of model scale and hint level. 
\cref{fig:iso_accuracy} shows that surface directly. The lines indicate 
iso-accuracy contours, along which accuracy is constant. 
Reading along a contour gives an exchange rate between 
hint level and model scale. 
The inverse slope of the contour gives the amount
of scale one unit of $H$ substitutes for at the contour's accuracy. 
For masking and truncation on both model families the contours run
as roughly parallel diagonals. This constitutes a very simple joint structure 
in which hint level trades linearly for model scale 
at all accuracies.

\section{Predicting Benchmark Performance}
\label{sec:benchmark_performance}

Having established that accuracy 
varies predictably with hint level and 
model scale, we turn 
to 
predicting the performance 
of scaled-up models. 
We adopt the model 
families and benchmarks of \cref{sec:phenomena}.
Our goal is to 
predict the accuracy of a family's largest members on the 
\emph{unhinted} benchmark using measurements on smaller 
members. 
Concretely, we fit a single 
law $L(C,H)$ to all hint levels of the $n$ smallest 
members of a family and evaluate its $H = 0$ slice on the 
held-out larger members.  

\textbf{Scaling law.}~~~
To fit a scaling law $L(C,H)$ 
we must select a functional form for $L$.
We can think
of a \method{} method and a functional form as a good pairing if two
conditions hold. The first is that $L$ has \emph{low fit error}. A single
law $L(C, H)$ must fit data across all scales. 
The second is that $L$ has
\emph{low extrapolation error} when fit on small-scale model data. Fitting on small-scale model data
with $H \neq 0$ included must identify the parameters of $L$ well enough to predict the upscaled target.

Thus we want a functional form for $L$ 
 that fits the data from 
\cref{sec:phenomena} well across \method{} methods 
and model families.
From \cref{sec:phenomena} we saw that accuracy was well modeled
as a sigmoid of hint level and model scale, with hint slices
whose asymptotes need not be 0 and 1 (\cref{fig:phenomena_ladders}).
This suggests that $L(C,H)$ could follow a sigmoid of
the form:
\begin{equation}
\label{eq:joint_law}
L(C, H) = \max\!\Big(F,\; L + (A - L)\,
\sigma\!\left(\beta_1 C + \beta_2 H + \beta_3\right)\Big),
\qquad
\sigma(z) = \frac{1}{1 + e^{-z}},
\end{equation}
where $C$ is either $\log_{10}$
FLOPs or $\log_{10}$ parameter count, and $F, L, A, \beta_1, \beta_2, \beta_3$ are
the fitted
parameters.
The upper asymptote $A$ lets accuracy saturate below 1.
The lower asymptote $L$ and the floor $F$ let accuracy sit flat
at $F$ until an emergence point, after which it follows a sigmoid.
We call \cref{eq:joint_law} the \emph{floored sigmoid law}.
With $A = 1$ and $L = F = 0$ it reduces to the plain sigmoid
$\sigma(\beta_1 C + \beta_2 H + \beta_3)$.
Additionally, this law suggests that
iso-accuracy curves of \cref{fig:iso_accuracy} above the floor have
the same slope $-\beta_1 / \beta_2$ everywhere.
From \cref{fig:iso_accuracy} we can see this is 
more reasonable for some family and method 
pairs than for others.
Most importantly, we find that in 
practice this law fits the data 
well
(\cref{tab:posttraining-well-spec}).

\textbf{Methods and data.}~~~~We use the model families and benchmarks  of 
\cref{sec:phenomena}.
Not every family makes for an informative forecasting 
problem on every benchmark, so we show a subset in the 
main body for brevity (with all pairs shown in \cref{app:benchmark_performance}).
In particular we exclude pairs if (1) every model in the ladder scores 0 
on the $H=0$ benchmark (making prediction trivial), or 
(2) every model in the ladder scores above 0 on the $H=0$
benchmark. From this, 
we report Qwen2.5 and Gemma-3 on AIME, Qwen3 and Qwen3.5 
on HLE, and Delphi on GSM8K.

We split our analysis up by the 
requirements of the \method{} method.
In \cref{sec:mask-trunc} we 
explore masking and truncation, which do
not require any teacher or access to model logits, for 
all families 
and benchmarks.
We show these methods are effective throughout.
In \cref{sec:logit-mixing} 
we explore logit mixing,
which requires a teacher model and access to model logits. We focus on the Delphi ladder and use Llama-3.1-8B-Instruct prompted with the solution in its context
as the teacher.

\begin{figure}[t]
    \centering
    \includegraphics[width=1\linewidth]{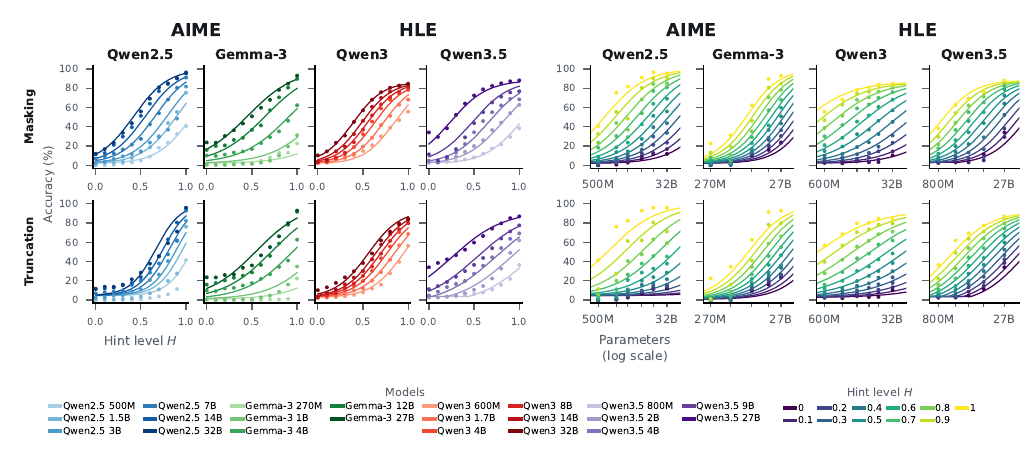}
    \vspace{-8mm}
    \caption{\textbf{\Method{} leads to scaling laws with low fit error.}
  Joint laws in $C$ and $H$ fit to the post-trained model families.
  Rows are the two \method{} methods, masking and truncation, columns the model family and benchmark.
  For each family and method we fit a \emph{single joint law}, \cref{eq:joint_law},
  to the accuracy of every model in the family at all eleven hint levels.
  \textbf{Left:} the hint slices (fixing model scale and letting 
  hint level vary) of each joint law. \textbf{Right:} the capability 
  slices (fixing hint level and letting model scale vary) 
  of each joint law.
}
    \label{fig:posttraining-well-spec}
\end{figure}

\subsection{A Simple Method Across Model Families and Benchmarks}
\label{sec:mask-trunc}

\begin{wraptable}{r}{0.46\linewidth}
  \vspace{-8mm}
  \centering
  \small
  \caption{Mean absolute error in percentage points of the fitted laws over the families of
  \cref{fig:posttraining-well-spec,fig:gsm8k-delphi-downscaling,fig:gsm8k-delphi-summing}.
  The joint law has six parameters per family, the individual fit averages the errors of
  each cell's hint- and capability-slice fits ($4|M| + 2|H|$ parameters). Logit mixing is
  Delphi only.}
  \label{tab:posttraining-well-spec}
  \vspace{1ex}
  \begin{tabular}{lcc}
    \toprule
    \textbf{Method} & \textbf{Joint Fit}  & \textbf{Individual Fit} \\
    \midrule
    Masking         & 3.1 & 1.8 \\
    Truncation      & 3.3 & 1.6 \\
    Logit mixing    & 1.7 & 1.7 \\
    \bottomrule
  \end{tabular}
\end{wraptable}

\begin{figure}[t]
      \centering
      \includegraphics[width=\textwidth]{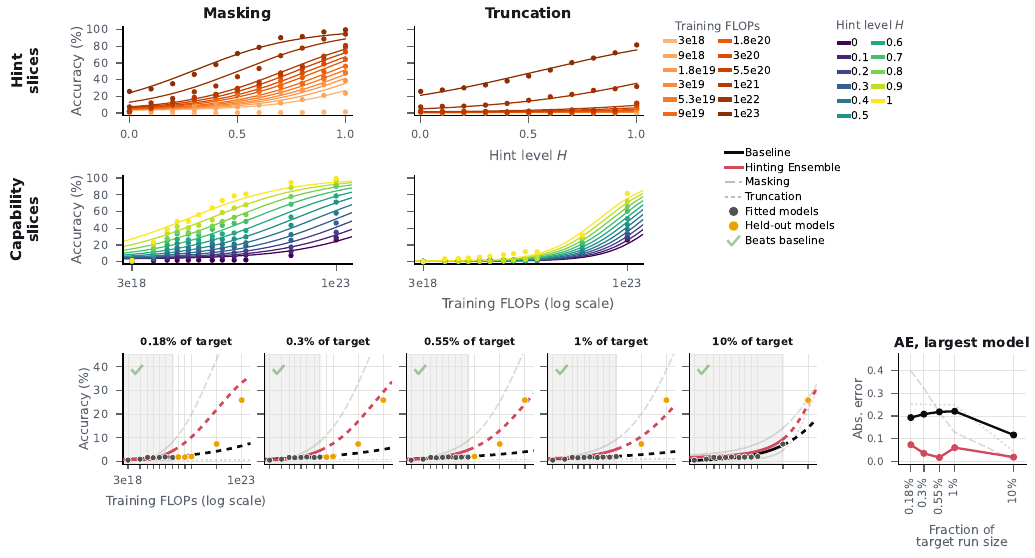}
      \vspace{-6mm}
     \caption{\textbf{\Method{} on the Delphi ladder (GSM8K).}
    \textbf{Top:} for each hinting method (masking and truncation) we fit one joint law
    (\cref{eq:joint_law}) over all models and 
    hint levels, with $C$ the $\log_{10}$ pretraining FLOPs. \textbf{Bottom:} the same law is refitted on only the
    $n$ smallest models and its unhinted ($H{=}0$) slice is evaluated at
    the larger, held-out models.
    The \textbf{black curve} is the baseline which fits a sigmoid to  unhinted accuracies alone. 
    Each individual method's forecast is
    shown in \textbf{\textcolor{methcol}{faint grey}}, with the
    \textbf{\textcolor{meancol}{red curve}} being the hinting ensemble forecast, the average
    of the masking and truncation forecasts. A \wintick marks panels in which the hinting ensemble
    forecast beats the baseline forecast on the largest ladder model. The right-most panel reports the absolute error (AE) on the 0--1 accuracy scale against the size of the largest training model.
    }
      \label{fig:gsm8k-delphi-downscaling}
  \end{figure}

\begin{figure}[!t]
    \centering
    \includegraphics[width=1\linewidth]{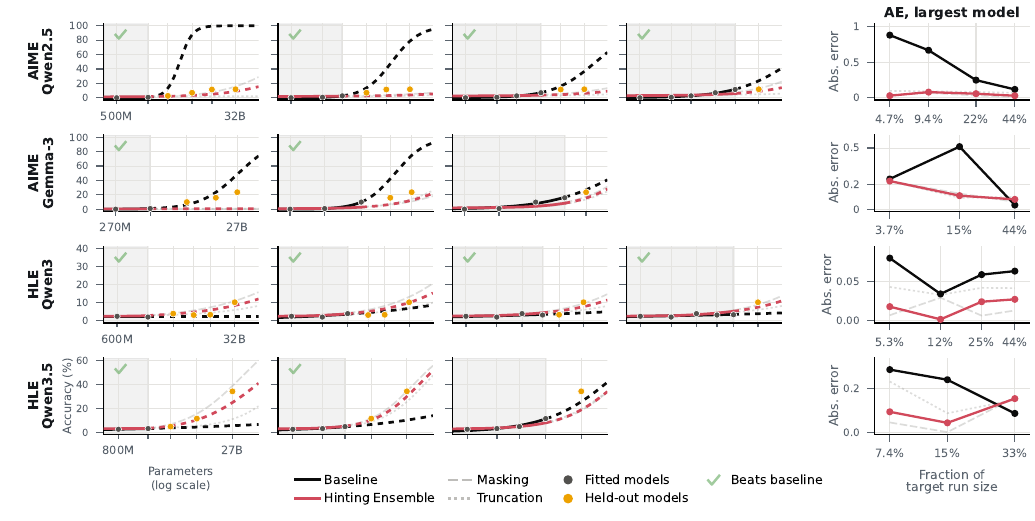}
    \vspace{-8mm}
    \caption{
\textbf{\Method{} leads to scaling laws with low extrapolation error.}
\Method{} forecasts of benchmark 
accuracy on post-trained model families.
Columns add train set models from left to right. For truncation 
and masking, we fit the 
joint law (\cref{eq:joint_law}) to all hint levels of the train set models, and use the $H = 0$ slice for the extrapolated curve. 
Each method is shown in \textbf{\textcolor{methcol}{faint grey}}, with the \textbf{\textcolor{meancol}{red curve}} 
being the hinting ensemble forecast, 
the average of masking and truncation 
forecasts. 
The \textbf{black 
curve} is the baseline sigmoid fit to original accuracies. 
A \wintick
marks panels in which our average forecast beats the baseline on the
largest model in the family. 
The right-most column reports absolute error (AE) on the 0--1 accuracy scale
against the size of the largest train set model.
}
    \label{fig:postraining-identifiable}
\end{figure}

Here we test masking and truncation across all model families and
benchmarks. These are the easiest \method{} methods to apply, as they only
require a reference solution and a change to the evaluated model's prompt.
We combine their forecasts with a simple \emph{hinting ensemble}. We fit a
separate law $L_{\omega}(C, H)$ for each
$\omega \in \{\text{truncation}, \text{masking}\}$ and predict with their
mean, $L(C, 0) = \frac{1}{2}\sum_{\omega} L_{\omega}(C, 0)$.
This also denoises our forecasts when a family has few models, as is
the case for the
post-trained families. We report both the ensemble forecast and the
individual method forecasts in
\cref{fig:postraining-identifiable,fig:gsm8k-delphi-downscaling}.

\textbf{\Method{} leads to scaling laws with low fit error.}~~~~In \cref{fig:posttraining-well-spec,fig:gsm8k-delphi-downscaling} 
we demonstrate that the joint 
law in \cref{eq:joint_law} has low fit error for truncation 
and masking. 
For each benchmark, \method{} method, and model family, we fit \cref{eq:joint_law} to every model's accuracy at eleven
hint levels, $H \in \{0, 0.1, \dots, 1\}$.
We draw slices
of that
single fitted surface at fixed hint and model scale. 
\cref{fig:posttraining-well-spec}
shows the results for the post-trained families,
and \cref{fig:gsm8k-delphi-downscaling} top for 
the pretrained models.

Fitting every slice on its own as we did in 
\cref{fig:phenomena_ladders} requires 
$4|M| + 2|H|$ parameters, where $|M|$ is the number of models in the family and 
$|H|$ is the number of hint levels. 
This is between 42 and 102 for our 
families, in comparison to 
the six parameters for each joint fit.
We compare the MAE (averaging 
over all models) of the joint 
fit against the
individual slice fits in \cref{tab:posttraining-well-spec}
and find that the joint fit, despite 
using between 36 and 96 fewer parameters, 
increases MAE by at most 1.7 percentage points. \cref{app:benchmark_performance} gives a second view of the same 
point, drawing the iso-accuracy contours the fitted joint law implies 
next to the empirical ones of \cref{fig:iso_accuracy}. The 
implied contours track the empirical ones.

\textbf{\Method{} leads to scaling laws with low extrapolation error.}~~~~In \cref{fig:postraining-identifiable,fig:gsm8k-delphi-downscaling} we demonstrate that the joint 
law for truncation and masking 
has low extrapolation error when fit on a small number of models. 
We fit the joint law of \cref{eq:joint_law} to all eleven hint levels of the $n$
smallest members of a family, and evaluate its $H = 0$ slice on the
larger members. The baseline is a simple two-parameter sigmoid 
fit to the original benchmark accuracy on the same $n$
smallest models. 
Both
methods therefore see the same models and the same unhinted scores, and
differ only in whether the hinted scores are available.
\cref{fig:postraining-identifiable} shows 
post-trained model results, and \cref{fig:gsm8k-delphi-downscaling} pretrained.

In many cases the baseline fails because 
it is not identifiable from the fitted models. 
When
the $n$ smallest members 
of a family sit near zero on the unhinted benchmark,
the fitted sigmoid is close to unconstrained leading to very poor 
extrapolation. On Qwen2.5 with two fitted models it 
predicts accuracy close to $100\%$ at 32B where the true value is $12\%$, an
absolute error of 88.2 percentage points. Our
average forecast from the same two models is off by 3.0 percentage points.  
Across all nineteen panels our average forecast beats the
baseline on the largest model in seventeen. We also compare 
to pass@$k$ as a hinting method (\cref{app:pass_at_k}), and a trivial hinting baseline 
that randomly gives 
the correct answer
to the model with probability $H$ (\cref{app:unassisted_determined}),
and outperform both.
Overall we see that hinted measurements allow us to identify 
the benchmark accuracy scaling law in many cases where 
unhinted measurements alone do not.

\subsection{Leveraging a Stronger Teacher}
\label{sec:logit-mixing}

\begin{figure}[t]
      \centering
      \includegraphics[width=\textwidth]{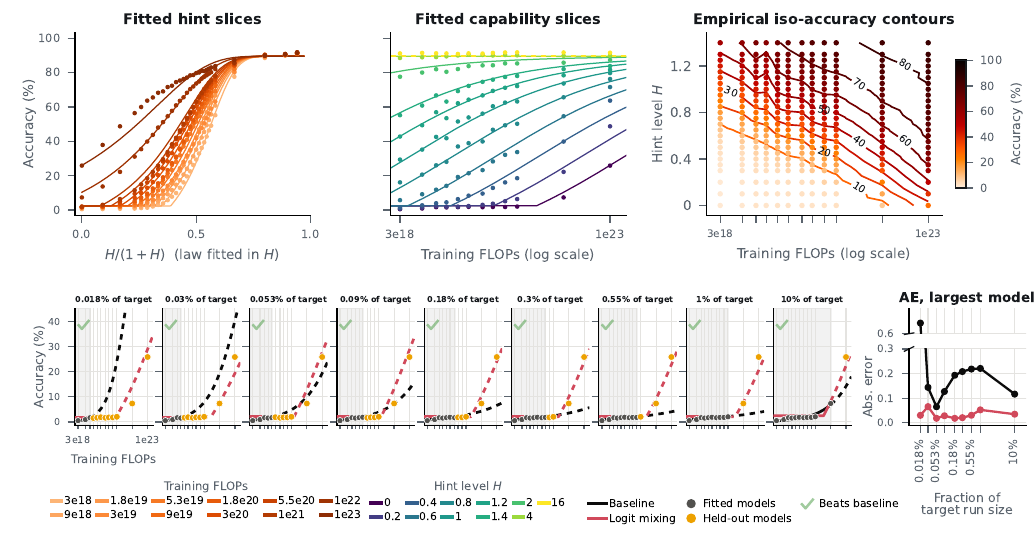}
      \vspace{-6mm}
     \caption{
     \textbf{Logit mixing with a shared teacher on the Delphi ladder (GSM8K).}
    \textbf{Top left and middle:}  Joint law fit
     over all models and 
    hint levels, with $C$ the $\log_{10}$ pretraining FLOPs. 
    \textbf{Top right:} empirical iso-accuracy contours of the same
    measurements for $H \leq 1.4$, as in \cref{fig:iso_accuracy}. 
    \textbf{Bottom:} the same law is refitted on only the
    $n$ smallest models and its unhinted ($H{=}0$) slice is evaluated at
    the larger, held-out models.
    The \textbf{black curve} is the baseline which fits a sigmoid to  unhinted accuracies alone. The \textbf{\textcolor{meancol}{red curve}} shows 
    our prediction. 
 }

      \label{fig:gsm8k-delphi-summing}
  \end{figure}

Masking and truncation reveal a partial solution to the evaluated model, so they
help only as much as the model can use the provided information. Delphi models
below $10^{22}$ FLOPs score at most 12\% under truncation even when shown the
whole solution ($H = 1$). Although ensembling the forecasts from masking and truncation does well, the two methods alone respectively under-forecast and over-forecast by a substantial amount when extrapolating beyond 10x compute.

Logit mixing steers decoding more directly using a teacher, so it can provide greater uplift for weaker models.
However, it requires a good teacher model that shares a tokenizer with the evaluated model to work well.
Here, we apply logit mixing to the Delphi ladder and instantiate the teacher by providing Llama-3.1-8B-Instruct with the full reference solution in its context (see \cref{app:logit_mixing} for implementation details).

Recall $H \in [0, \infty)$ for logit mixing. In 
our experiments, we sweep $H$ values between 
$0$ and $16$. In \Cref{fig:gsm8k-delphi-summing} 
we show the results of fitting 
\cref{eq:joint_law} in $H$ and using it to 
predict scaled up model performance.
Firstly, we see that 
a shared teacher changes the shape of hint slices. However 
weak the evaluated model, raising $H$ pulls its 
next-token distribution toward the
teacher, so every hint slice rises to a common ceiling set by the
teacher. The upper asymptote $A$
of \cref{eq:joint_law} captures this ceiling directly.  
The iso-accuracy contours (\cref{fig:gsm8k-delphi-summing}, top right)
are also simpler than those of masking and truncation in
\cref{fig:iso_accuracy}. They run as nearly parallel straight lines, which
is the structure \cref{eq:joint_law} assumes above its floor, with every
contour at slope $-\beta_1/\beta_2$.

Fitted to every model and hint level, the joint law has a very low mean
absolute error of 1.7 percentage points 
(\cref{fig:gsm8k-delphi-summing},
top), the same as fitting every slice on its own
(\cref{tab:posttraining-well-spec}). The fitted floor $F = 2.5\%$ sits just above the 0.5 to 2\% unhinted
accuracy of every model below $10^{22}$ FLOPs. 
The mean forecasting error across all nine training sets is 3.3 points, and the largest error of 6.6 points occurs when extrapolating from 0.03\% of the target compute budget. The forecast stabilizes at around 0.2\% of the target compute budget ($10^{23}$ FLOPs).

The original Delphi ladder consisted of 10 compute budgets and a single random seed for every budget except $10^{21}$ and $10^{22}$ FLOPs. We trained two additional random seeds for each of the 7 smallest original budgets as well as three seeds at new budgets of $5.3 \times 10^{19}$ and $5.5 \times 10^{20}$ FLOPs. We fit our joint law after averaging over the hinted accuracies for all available seeds at each compute budget (we have only one seed available for $10^{23}$ FLOPs).
We find using multiple random seeds is 
important: forecasting with only the original seed at each budget obtains a mean error of 10.2 points over all nine training sets.

For the post-trained model families, there does not always exist a strong external model with the same tokenizer.
We include experiments in \cref{app:logit_mixing_posttrained} that instantiate the teacher model for each post-trained family by prompting a model from within the family (i.e., the largest model in the training set we use for the forecast) with the solution in its context. We find that although these teachers do achieve high benchmark accuracy, they tend to produce somewhat unnatural solutions (e.g., they explicitly reference the ``hint'' in their generations) and produce poor forecasts.
We did not meaningfully explore alternative instantiations of the teachers for the post-trained families (e.g., via prompting or finetuning teachers to produce more natural solutions). We also did not meaningfully explore logit mixing variants \citep{hayase2026sampling,xu2024bridging,chawla2026alignbeam} that do not require the teacher to share a tokenizer with the evaluated models.
We consider these to both be promising directions for future work.

Finally, a quirk of logit mixing is that the accuracy improves at moderate to high hint levels if we assign zero weight to the evaluated model's logits (or equivalently, rescale its logits to have infinite temperature). This implies that the capability slices for logit mixing are not necessarily monotonic: a weak enough model whose output distribution is close to uniform may achieve higher accuracy than stronger models for a high enough hint level, particularly with an overly confident teacher (e.g., a teacher that overfits to the in-context solution).
We discuss this issue in more detail in \cref{app:logit_mixing_posttrained} along with some potential remedies and alternative methods.

\section{Ranking Models by Benchmark Accuracy}
\label{sec:model_ranking}

Next, we explore whether hinting can help with sample-efficient model ranking.
In this task, we are given a set of models with similar expected 
accuracy on a benchmark and we wish to find the true ranking 
of expected accuracies of the models with as few model 
rollouts as possible.
We focus on models with similar accuracies near zero, where tiny accuracy gaps make an unhinted ranking need many rollouts. Hinting widens these gaps while approximately retaining the rank order so that fewer rollouts are needed.
While \cref{sec:benchmark_performance} tested whether hinting can help predict performance of models across a wide range of capabilities, here we test whether hinting can tell apart models of nearly identical capability.

\textbf{Setting.}~~~~We call models \emph{near zero} if all their accuracies lie within a few percentage points of the floor and \emph{nearly indistinguishable} if their accuracies span a very small range. We build one such set of seven models for AIME and another for HLE. Their unhinted accuracies span $0.06\%$ to $0.81\%$ on AIME and $1.7\%$ to $3.0\%$ on HLE. We obtain a ground-truth ranking from 4{,}096 and 1{,}024 unhinted rollouts per problem respectively, drawn separately from the rollouts the ranking methods use 
(see \cref{app:ranking_ground_truth}). We report results for the truncation hinting method in the main text, which we find works best for ranking (see \cref{app:ranking_other_methods} for other methods).
We evaluate a ranking by Kendall's $\tau$ \citep{kendalltau} with the ground-truth ranking, where $\tau = 1$ is perfect agreement and $\tau = 0$ is chance. Each ranking method receives a budget of $k$ rollouts per problem per model, and we report $\tau$ averaged over 1{,}000 simulated runs at that budget.

We compare three rankings: \emph{avg@$k$}, which averages the $k$ unhinted rollouts; a \emph{log-loss baseline}, which needs no rollouts and ranks models by their teacher-forced log loss on the reference solutions; and our \emph{hint-prior ranking}. The hint-prior ranking spends one rollout per problem on a hinted benchmark $\bar H$ that mixes hint levels across problems and the remaining $k - 1$ on the unhinted one. 
Writing $\widehat{\mathrm{Acc}}_1(M, \bar H)$ and $\widehat{\mathrm{Acc}}_{k-1}(M, 0)$ for the resulting hinted and unhinted scores and $z[\cdot]$ for the z-score across models, it ranks models by

\begin{equation}
\label{eq:assistance_prior}
\hat{s}(M, k)
= (1 - \alpha_k)\, z\!\left[\widehat{\mathrm{Acc}}_1(M, \bar H)\right]
+ \alpha_k\, z\!\left[\widehat{\mathrm{Acc}}_{k-1}(M, 0)\right],
\qquad \alpha_k = \frac{k - 1}{k - 1 + \lambda},
\end{equation}

so for $k \leq 1$ the ranking uses the hinted score alone and the weight shifts to the unhinted score as $k$ grows. We choose this because the hinted score has low variance but a bias that does not shrink with $k$, whereas the unhinted score is unbiased with variance that does.
Another design would be to fit a sigmoid over hint level per model and rank by its value at $H = 0$, but this needs many rollouts at $H > 0$, which at a fixed budget might be better spent on a single hinted rollout per problem and on the unhinted accuracy itself.
We set $\lambda = 8$ throughout, so the hinted score counts as much as eight unhinted rollouts per problem. \cref{app:ranking_hint_prior} gives the construction and the choice of $\lambda$.

\begin{figure}[t]
    \centering
    \includegraphics[width=1\linewidth]{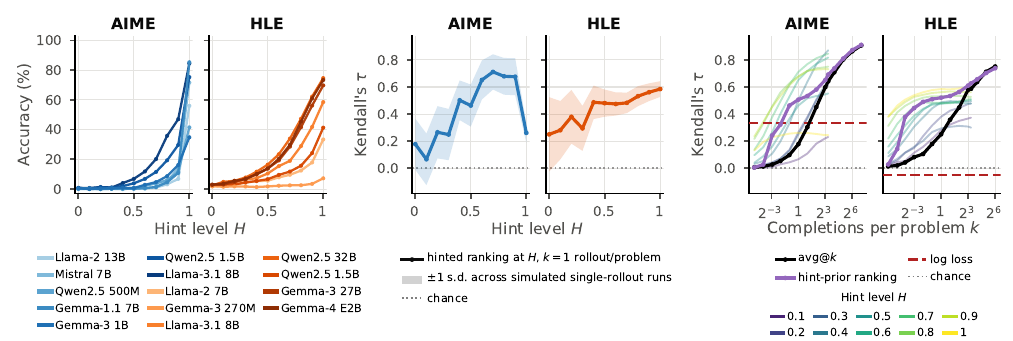}
    \vspace{-6mm}
    \caption{\textbf{\Method{} largely preserves the model ranking and recovers it from fewer rollouts} (truncation hinting; masking in \cref{fig:model_ranking_all_methods}).
  \textbf{Left:} accuracy versus hint level $H$, one curve per model, colored by ground-truth rank at $H = 0$ from light (lowest) to dark (highest).
  \textbf{Middle:} Kendall's $\tau$ between the ground-truth ranking and the ranking from one rollout per problem at each $H$. Bands are $\pm 1$ std over simulated single-rollout runs.
  \textbf{Right:} Kendall's $\tau$ against the budget of $k$ rollouts per problem for \textbf{\textcolor{avgkcol}{avg@$k$}}, the \textbf{\textcolor{hintshrinkcol}{hint-prior ranking}} (\cref{eq:assistance_prior}), the single-level rankings of the middle panel (\hintgradient{}), and the \textbf{\textcolor{loglosscol}{log-loss baseline}}.}
    \label{fig:model_ranking}
\end{figure}

\textbf{Hinting preserves ranking.}~~~~In \cref{fig:model_ranking} (left), hinting spreads the models' accuracies apart while largely preserving their unhinted order. Darker curves indicate higher performance on the original benchmark ($H = 0$), and we observe that with increasing $H$, curves roughly stay in color order. Not every hint level preserves it equally well, however. \cref{fig:model_ranking} (middle) shows how well one hinted rollout per problem recovers the ranking at each level, and the best level differs between benchmarks: Kendall's $\tau$ peaks at $H = 0.7$ on AIME and drops sharply at $H = 1$, while on HLE it rises through $H = 1$. The hint-prior ranking avoids tuning $H$ by using all hint levels.

\textbf{\Method{} enables sample-efficient estimation of the ground-truth ranking.}~~~~At small budgets, the hint-prior ranking is far closer to the ground truth than avg@$k$, the ranking by unhinted accuracy alone (\cref{fig:model_ranking}, right). At one rollout per problem, it leads avg@$k$ by $0.33$ in Kendall's $\tau$ on AIME and $0.27$ on HLE, and avg@$k$ requires about $5$ rollouts per problem on AIME and $6$ on HLE to match it. 
The lead narrows as the budget grows and the weight shifts to the unhinted score. 
The two converge at $128$ rollouts per problem on AIME and $64$ on HLE, the largest budgets we evaluate, and at no budget is the hint-prior ranking behind avg@$k$ by more than $0.02$ in $\tau$. 
This convergence holds by construction: $\alpha_k$ approaches one as the budget grows, so the hint-prior ranking becomes avg@$k$ at large budgets whether or not the hinted ranking is correct.
The log-loss baseline uses a continuous metric rather than averaging rollouts, and does worse than both sampling methods.
It reaches $\tau = 0.33$ on AIME, a level avg@$k$ needs about two rollouts per problem to match and the hint-prior ranking below one, and $\tau = -0.05$ on HLE, below chance.

\textbf{Near zero orderings depend on the metric.}~~~~A subtle nuance in our setting is that it is not clear what the target ranking should be. An alternative to avg@$k$ for measuring the capability of weak models is pass@$k$, and these metrics do not necessarily agree on the ranking. On AIME, pass@$k$ ranks the seven models in the same order as the ground truth average up to $k = 50$. On HLE it does not: the agreement falls from $\tau = 0.62$ at $k = 5$ to $0.05$ at $k = 50$.
Both of these metrics putatively measure model capability yet their rankings disagree, implying there is no single ordering that a measure of capability could recover. 
This could partially explain why our ranking methods do worse on HLE than on AIME.

\section{Related Work}
\textbf{Downstream scaling laws.}~~~~Downstream scaling laws extrapolate benchmark accuracy from a ladder of small models, but on challenging benchmarks these laws can be unreliable, as these models may score close to chance which leaves the fit unconstrained~\citep{lourie2025scalinglawsunreliabledownstream}.
Prior work gets around this in one of three ways.
The first approach fits the law only on models that already score above chance, either directly~\citep{krajewski2025revisitingscalingpropertiesdownstream} or through a loss proxy~\citep{chen2025scalinglawspredictingdownstream, bhagia2025establishingtaskscalinglaws, gadre2024languagemodelsscalereliably, grattafiori2024llama3herdmodels}. \citet{held2026delphi} go further and forecast the accuracy of the largest Delphi model by fitting the loss-to-accuracy step on external models, since the whole ladder is at chance.
The second approach keeps models at chance but replaces accuracy with a continuous metric such as the likelihood of the correct answer~\citep{schaeffer2023emergentabilitieslargelanguage, magnusson2025datadecidepredictbestpretraining, olmo2026olmo3, koh2026predictingllmreasoningperformance, heineman2025signalnoiseframeworkreducing}, which has signal at small scale but is not the metric of interest.
The third approach keeps both the at-chance models and the metric, but pays extra computational cost to sample each problem until the model succeeds~\citep{hu2024predictingemergentabilitiesinfinite} or $k$ times and fit a law to pass@$k$~\citep{krajewski2025revisitingscalingpropertiesdownstream}. The downside to this approach is that increasing the signal in this way incurs significant (and sometimes intractable) computational costs~\citep{brown2024largelanguagemonkeysscaling, schaeffer2025largelanguagemonkeyspower}.
\citet{snell2024predictingemergentcapabilitiesfinetuning} instead finetune the model family on varying amounts of task data and fit a law in that amount. %
We introduce a unified framework for such methods which vary evaluation along a second parameter (number of rollouts, amount of finetuning data) but still predict accuracy on the original benchmark.
Outside language modeling, \citet{jones2021scalingscalinglawsboard} scales the difficulty of board games and finds that the compute--performance curve keeps the same sigmoid shape across board sizes, only shifted. Our joint law plays the same role with hint level in place of board size.

\textbf{Sample-efficient model ranking.}~~~~Prior work ranks models with less evaluation by scoring only a subset of problems~\citep{zhang2026benchmarkpredictionfewerdata, polo2024tinybenchmarksevaluatingllmsfewer, hofmann2025fluidlanguagemodelbenchmarking, balkir2026confidentrankingsfeweritems}, but no subset of problems reliably orders models fewer than a few accuracy points apart~\citep{yauney2026reliablelanguagemodelmicrobenchmarking}. Others replace generation with the likelihood of a reference solution~\citep{olmo2026olmo3, patel2026forecastingdownstreamperformancellms}. We rank from few rollouts per problem by using hinted accuracy as a prior (\cref{sec:model_ranking}).

\textbf{Hinting in prior work.}~~~~Closest to our truncation method,
\citet{li2025stayingsweetspotresponsive}, \citet{qu2026popelearningreasonhard} and \citet{setlur2026reuseflopsscalingrl}, and concurrently \citet{agrawal2026offcontextgrpolearningreason}, condition reinforcement learning rollouts on a prefix of a correct solution so that problems the model never solves still yield a learning signal. At evaluation time, \citet{zhou2025gradersshouldcheatprivileged} give candidate models hints created from the reference solution in discrete tiers, which makes near zero models distinguishable.
Concurrently with our work, \citet{yu2026solvepredictingposttrainingcodingagent} rank base checkpoints with near zero accuracy by conditioning them on a prefix of a successful agent trajectory. We make the hint level continuous and fit a law across it. \citet{wen2026pricedguidancelanguagemodels} let a model ask a guide that knows the answer for information, charging each answer its information content in bits under the model's own prior, which suggests a new hinting method where the amount of hint is a bit budget.

\section{Discussion}

This work initiates the study of hinted scaling laws and posits that jointly modeling model capability and hint level enables better predictions of benchmark accuracy than modeling capability alone. 
We consider two prediction problems as applications: forecasting benchmark accuracy across a model family and ranking models from different families.

The main limitation of our forecasting study is the scarcity of open-source scaling ladders. Among the five model families we consider, only Delphi is explicitly a scaling ladder trained on one data recipe. Thus, the extent to which our forecasting errors are attributable to shortcomings of our methods versus artifacts of these model families is unclear. Certainly, our simple joint law is an imperfect model of the true joint effect of hint level and model capability on accuracy.

Generalizing our work beyond what we presented, developing sample-efficient methods for ranking LLM agents on benchmarks is particularly important in view of the substantial cost of running modern safety-relevant agentic benchmarks. For example, a single run of UK AISI's cyber-range evaluation costs roughly \$3000--\$5000 per rollout per frontier model \citep{aisi2026cyber}, and safety and red-teaming benchmarks need many rollouts to estimate long-tail risks \citep{angell2026estimating,jones2025forecasting,ICLR2025_cdaac2a0}.%
\footnote{The cyber-range evaluation caps models to 100M tokens per rollout. Current prices for GPT 5.5 (\$30/1M tokens) and Claude Mythos priced as Claude Fable (\$50/1M) then lead to costs of respectively \$3,000 and \$5,000 per evaluation iteration.}
Our ranking experiments use single-turn benchmarks, but hinting extends naturally to agentic benchmarks where a hint can replay part of a successful trajectory and let the model finish the task from that state. We do not explore hinting for safety and red-teaming evaluations in this paper, but we view this as a particularly worthwhile endeavor for future work.

Finally, both applications rely on benchmarks for which there exist verified solutions with legible reasoning for each problem. Given the rapidly expanding capability frontier of language models, we expect it will be increasingly difficult to construct benchmarks of this type that are not already saturated. Nonetheless, even in such a world, we expect hinted scaling laws will still be useful to model developers, as there will always be a need to estimate model performance from a small scale when optimizing the compute efficiency of a training recipe~\citep{hernandez2020measuringalgorithmicefficiencyneural}. Going beyond capabilities, hinting may also play an increasingly important role in helping model developers more precisely control complex and emergent behaviors such as reward hacking and multi-agent collusion.

\section*{Acknowledgments}

LB thanks the support of a Stanford Graduate
and Vitalik Buterin Fellowship.
RK acknowledges support from an NSF Graduate Research Fellowship.
NB acknowledges support from an NSF Graduate Research Fellowship, Quad Fellowship, and Mercor Graduate Fellowship.
SV was supported by CURIS, Stanford's undergraduate Computer Science research program.
PL was supported by the DARPA AIQ program.
TH was
supported by a grant by HAI, DSO labs, 
gifts from Open Philanthropy, Amazon, 
Schmidt Sciences,
the Tianqiao and Chrissy Chen Foundation
and a grant under the NSF
CAREER IIS-2338866, ONR
N00014-24-1-2609, and DARPA Cooperative 
Agreement HR00112520013. This work does not
necessarily reflect the position or policy of the government and no official endorsement should be
inferred.

We acknowledge compute support from Google TPU Research Cloud and the Marin Project.

We thank Tony Lee for help with early experiments and Romain Yon for support maintaining infrastructure for inference on TPU chips.
We thank Lars Ankile, Elise Han, Will Held, Tanishq Kumar, and Kevin Xiang Li
for helpful feedback during the
project. We thank all members of 
the Hashimoto group and P$\lambda$ for their feedback.
We thank Marka Ellertson and Louise van Adrichem for 
help with writing.

\newpage

\ificlr
  \bibliographystyle{iclr2027_conference}
\else
  \bibliographystyle{plainnat}
\fi
\bibliography{refs}

\appendix
\ificlr\else
  \renewcommand{\topfraction}{0.95}
  \renewcommand{\textfraction}{0.05}
  \renewcommand{\floatpagefraction}{0.9}
\fi
\newpage
\section{Benchmark Details}
\label{app:benchmark_details}

\paragraph{AIME.} We use the 60 problems of the 2025 and 2026 AIME~I
and~II exams taken from MathArena. Every answer is an integer between
0 and 999. We prompt models to put their final answer in
\texttt{\textbackslash boxed\{\}}, extract the final answer from
the response with a rule-based parser, and mark the response correct
if it equals the official answer.

\paragraph{HLE.} From the 2{,}500 questions of Humanity's Last Exam we use a
176-question subset. First, we consider only questions without
images. Of these, we keep a question if (1) its answer is free-form
(\texttt{exactMatch}) rather than multiple-choice, (2) we can generate
a correct reference solution for it (\cref{app:reference_solutions}),
and (3) it does not hand the model a set of candidate answers. For (3)
we manually removed 14 free-form questions that ask a yes/no question,
enumerate options in the question text, or ask ``how many of the
following''.
Following the HLE protocol, we judge correctness with an LLM. We use
the HLE judge prompt with its confidence field removed, since we do not
ask models for a confidence score, and GPT-5.4 nano (low reasoning
effort) as the judge. The judge sees the question, the ground-truth
answer and the final 1{,}500 characters of the model's response, and
returns the extracted final answer and a yes/no verdict. 

\paragraph{GSM8K.} We use the first 256 problems of the GSM8K test
split. The reference solutions are the step-by-step solutions that
come with the benchmark, with the final answer line removed; for truncation we also replace any remaining occurrence of the answer with \texttt{<mask>}.
We grade with strict matching. We take the first number after the first ``\texttt{\#\#\#\#}'' in the response, drop dollar signs and commas, and cast the answer to a float (effectively dropping trailing decimal zeros). We then compare it to the ground truth. A response with no ``\texttt{\#\#\#\#}'' counts as incorrect.
We include ``Question:'', ``\textless/s\textgreater'', and ``\textless\textbar im\_end\textbar\textgreater{}'' as stop strings.

\paragraph{Sampling.} \Cref{tab:sampling} gives the sampling settings. Post-trained models use the settings their model cards recommend, and \cref{app:logit_mixing} gives the logit-mixing settings.

\begin{table}[h]
\caption{Sampling settings per model family. Gemma-1.1, Gemma-2 and Llama-2 generate up to their shorter context windows.}
\label{tab:sampling}
\begin{center}
\footnotesize
\setlength{\tabcolsep}{4pt}
\begin{tabular}{lccccl}
\toprule
\bf Family & \bf Temp. & \bf Top-$p$ & \bf Top-$k$ & \bf Token limit & \bf Other \\
\midrule
Qwen3, Qwen3.5 & 0.6 & 0.95 & 20 & 32{,}000 new & \\
Qwen2.5 & 0.7 & 0.8 & 20 & 32{,}000 new & rep.\ penalty 1.05 (1.1 for 0.5B, 1.5B) \\
Gemma-3, Gemma-4 & 1.0 & 0.95 & 64 & 32{,}000 new & \\
Gemma-1.1, Gemma-2 & 1.0 & 0.95 & 64 & 8{,}192 context & \\
Llama-3 & 0.6 & 0.9 & 50 & 32{,}000 new & \\
Llama-2 & 0.6 & 0.9 & 50 & 4{,}096 context & \\
Mistral & 0.7 & off & off & 32{,}000 new & \\
\midrule
Delphi, masking and truncation & 0.4 & 1 & 1000 & 512 new & 5-shot, no chat template \\
Delphi, unhinted & 0.4 & 1 & off & 512 new & 5-shot, no chat template \\
\bottomrule
\end{tabular}
\end{center}
\end{table}

\section{Reference Solutions for AIME and HLE}
\label{app:reference_solutions}

\Method{} methods require a correct solution to every problem. For example, both masking and truncation hinting give part of the solution to the model. Such a correct solution should not state the final answer; otherwise the model could copy it without building on the hint. This is especially relevant for masking, where the hint level $H$ would otherwise just be the probability of revealing the answer to the model. Thus, we make sure that the final answer is not in the correct solution. 

We build the solutions in two stages. First we sample full solutions from a frontier model and keep only ones that are deemed correct by a grader. 
Second, we make a model rewrite each correct solution so that it does not include the final answer and stops just before it.  We collect ten reference solutions per problem on both benchmarks, and each hinted rollout of a problem uses one of them.

\paragraph{Stage 1: sampling correct solutions.}
We prompt a frontier model with the problem and keep a sample only if it is graded
correct. On AIME a sample is correct if the extracted integer matches
the official answer; on HLE we apply the LLM judge of
\cref{app:benchmark_details} with o3-mini (low reasoning effort) as
the judge. We resample until every problem has ten correct solutions.
On AIME the solver is Claude Opus 4.6 with extended thinking off, or at
medium or high effort. The remaining rollouts of a few problems on
which Opus 4.6 rarely succeeded
were sampled with GPT-5.4 with medium thinking. On HLE the solver is
GPT-5.5 with medium thinking.

\paragraph{Stage 2: removing the answer.}
After collecting correct solutions, we make a model rewrite them so they do not contain the problem answer. This second model is given
the problem, the solution and the answer, and is instructed to stop
before the final step and to never state the answer. On AIME the
rewriter is Claude Opus 4.6 with extended thinking off. 
On HLE the rewriter is Claude Opus 4.7 with medium
thinking. 
The rewritten
solutions are the reference solutions used by masking and truncation on
AIME and HLE. For logit mixing, the teacher uses the benchmark's own solutions on GSM8K
and these rewritten solutions on AIME (\cref{app:logit_mixing_posttrained}).

\paragraph{Checking for leaked answers.}
To check that the rewritten solutions do not state the final answer, we search each for its problem's answer string with a regular expression. The string occurs in 19 of the 600 AIME solutions and 158 of the 1,760 HLE ones. On inspection, very few (only 4 and 2) of these state the answer; the rest contain it incidentally, as a quantity in the problem, an intermediate value, part of a larger number, or the value just before the final step.

\section{Logit Mixing Details}
\label{app:logit_mixing}

\cref{sec:methods} describes logit mixing over the full vocabulary. In
practice, we implement logit mixing by running two inference engines on separate devices and communicating only their 16 largest logits to a shared host (increasing the number of logits slows down generation).
At each decoding
step, we mix over the union of the two candidate sets. A candidate missing from one side receives that side's smallest reported logit. We then sample
$y_t \sim \mathrm{softmax}(s_H / T)$ over the union, with $s_H$ from
\cref{eq:logit_sum} and $T = 0.4$. The evaluated Delphi model receives the
standard 5-shot GSM8K prompt. The teacher, Llama-3.1-8B-Instruct, receives
a chat history in which the reference solution appears as its previous
answer followed by the request ``Great. Can you solve it another way?''
and an ``Answer:'' prefill, so that it continues with a fresh
solution.
We treat Llama-3.1-8B-Instruct's end-of-turn token as its stop token (i.e., we mix its logit with Delphi's stop token logit) and discard all other special tokens.
We sweep 29 values of $H$ from 0 to 16, denser below $H = 1.4$,
and draw 32 samples per problem on the first 256 GSM8K test problems for
each of the three seeds below $10^{23}$ FLOPs. For masking and truncation
on Delphi we sample at the same temperature with a top-$k$ of 1000, 32 rollouts per problem and seed. The
$H = 0$ points are unhinted runs shared by all three methods, 64 rollouts per problem and seed with top-$k$ off.

\subsection{Logit Mixing on Post-Trained Families}
\label{app:logit_mixing_posttrained}

Logit mixing needs a teacher that is more capable than the evaluated models
and shares their tokenizer. On the Delphi ladder such a teacher is readily
available. On the post-trained families the choice is less clear. Most
families have their own tokenizer, so the teacher must come from the same
family, and the natural candidates are its largest models. Using the forecast
target itself as the teacher would leak its accuracy into the forecast. We
therefore take the second-largest model of each family as the teacher, place
the reference solution in its context, and mix it with every model in the
family up to and including itself. We fit the joint law on every model except
the largest and forecast only the largest model's unhinted accuracy, so the
target is never used as a teacher or in the fit. We run these experiments on
AIME for Qwen2.5, Qwen3 and Gemma-3.

We note that this experimental setup \emph{restricts} our 
ability to evaluate logit mixing forecasts as rigorously 
as the other forecasting experiments in the paper. Namely 
we can only evaluate forecasting from all but the largest 
model in the family to the largest model. 
Logit mixing is most natural when a stronger external teacher outside the
family is available, as on the Delphi ladder in \cref{sec:logit-mixing}, and
we include these post-trained results for completeness.

\begin{figure}[t]
    \centering
    \includegraphics[width=\linewidth]{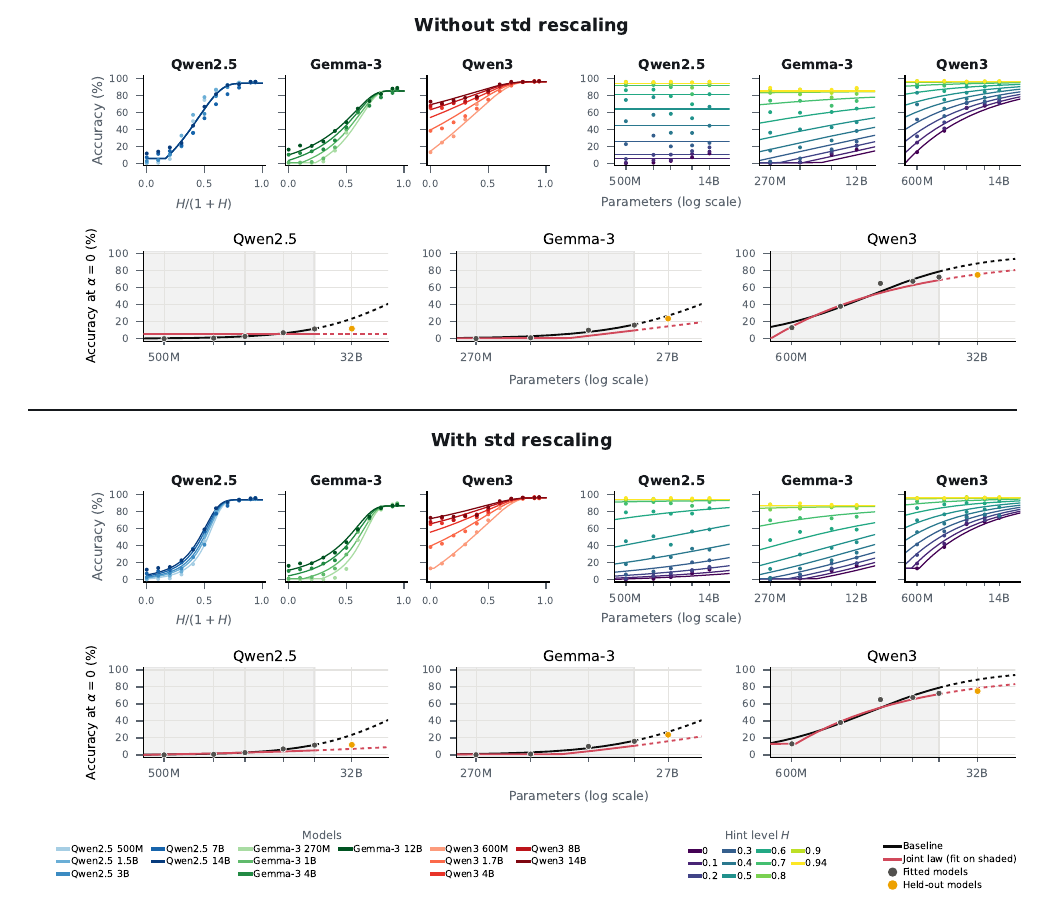}
    \caption{\textbf{Logit mixing on post-trained families 
    on AIME.}
    The top half uses the raw rule (\cref{eq:logit_sum}), and the bottom half
    rescales the teacher's logits to the evaluated model's standard deviation at
    each step (\cref{eq:logit_sum_std}). Each
    half shows the hint and capability marginals of the joint law fitted to
    every model except the largest, and its forecast of the largest model at $H = 0$.}
    \label{fig:logit_mixing_posttrained}
\end{figure}

\textbf{Raw logit mixing.}~~~~The top half of \cref{fig:logit_mixing_posttrained}
uses the rule of \cref{eq:logit_sum}. On Qwen3 and Gemma-3, accuracy rises
smoothly with both the hint level and model scale, and the joint law fits
both families. Qwen2.5 is the exception. At intermediate hint levels its
smaller models score higher than its larger ones. This 
behavior is completely opposite to what we want from hinting,
and accordingly the joint fit and forecast are very poor.

\textbf{Flat logits and std rescaling.}~~~~The inversion comes from how the
logit mixing raw rule weights the teacher. 
In \cref{eq:logit_sum} the teacher's ``effect'' at a
given $H$ depends on how confident the evaluated model is, and the 
scale of its logits. 
A student model whose
logits are flat and small in scale, when summed with even 
a small amount of teacher logits, ends up sampling from
the teacher more frequently than a student that has 
a sharp distribution that disagrees with the teacher.
This means that larger models can receive a 
``weaker'' hint at the same $H$.

On Qwen2.5, the gap between the evaluated model's
largest and 16th largest logit grows from 14.0 for the 0.5B model to 20.9
for the 7B model. At $H = 1$, on the steps where the two models' top tokens
differ, the mixed distribution's top token is the teacher's on 51\% of steps
for the 0.5B model but only 37\% for the 7B model. 

One simple fix is to
rescale the teacher's logits to the evaluated model's scale at each step,
\begin{equation}
\label{eq:logit_sum_std}
\begin{split}
s_H(x, y_{<t})_j &= z_{\mathrm{model}}(x, y_{<t})_j
  + H\, \frac{\sigma_{\mathrm{model}}}{\sigma_{\mathrm{teacher}}}\, z_{\mathrm{teacher}}(x, y_{<t})_j \\
  &= \sigma_{\mathrm{model}} \left( \frac{z_{\mathrm{model}}(x, y_{<t})_j}{\sigma_{\mathrm{model}}}
  + H\, \frac{z_{\mathrm{teacher}}(x, y_{<t})_j}{\sigma_{\mathrm{teacher}}} \right),
\end{split}
\end{equation}
where $\sigma$ is the standard deviation of a model's 16 largest logits at
that step, and all other symbols are 
as defined in \cref{eq:logit_sum}. We call this method 
std rescaling. We can interpret the above equation as putting the student and teacher 
logits on the same scale before adding, and then scaling back 
up to the student model scale. Note importantly that this 
preserves $H=0$ corresponding to sampling from the evaluated model,
and $H\to \infty$ sampling from the teacher.

The bottom half of
\cref{fig:logit_mixing_posttrained} shows the result. On Qwen2.5, accuracy
now largely increases with scale at every hint level, and the forecast error
for Qwen2.5-32B falls from 6.4 to 4.8 percentage points. The forecasts for
Qwen3 and Gemma-3 stay similar to their prior results.

We note that std rescaling is one of several possible corrections, which we
chose after observing the Qwen2.5 inversion, and we have tested it only on
AIME. We therefore present it as a possible fix for 
this pathology of logit mixing rather than
as part of the method.
Another possible correction is to sweep over rescaling the evaluated model's logits at each hint level and take the best performing rescaling factor; this ensures by construction that every model's hinted performance is at least as good as a model that samples tokens uniformly at random.
We piloted a coarse version of this correction on the Delphi ladder, but did not pursue the approach further due to the expensive nature of the sweep.
We also piloted other methods for leveraging the teacher model besides logit mixing, such as generating several candidate token spans from the student and selecting the best span using the teacher (this particular method was expensive to run, and we did not observe significant uplift for spans beyond a couple tokens in length). We leave more extensive exploration of logit mixing alternatives to future work.
We note that we 
did not observe any pathology for the logit mixing results in 
the main text that use the
raw rule of \cref{eq:logit_sum}.

\section{Hinting and Capability Scaling: All Families}
\label{app:all_families}

\begin{figure}[htbp]
    \centering
    \includegraphics[width=0.85\linewidth]{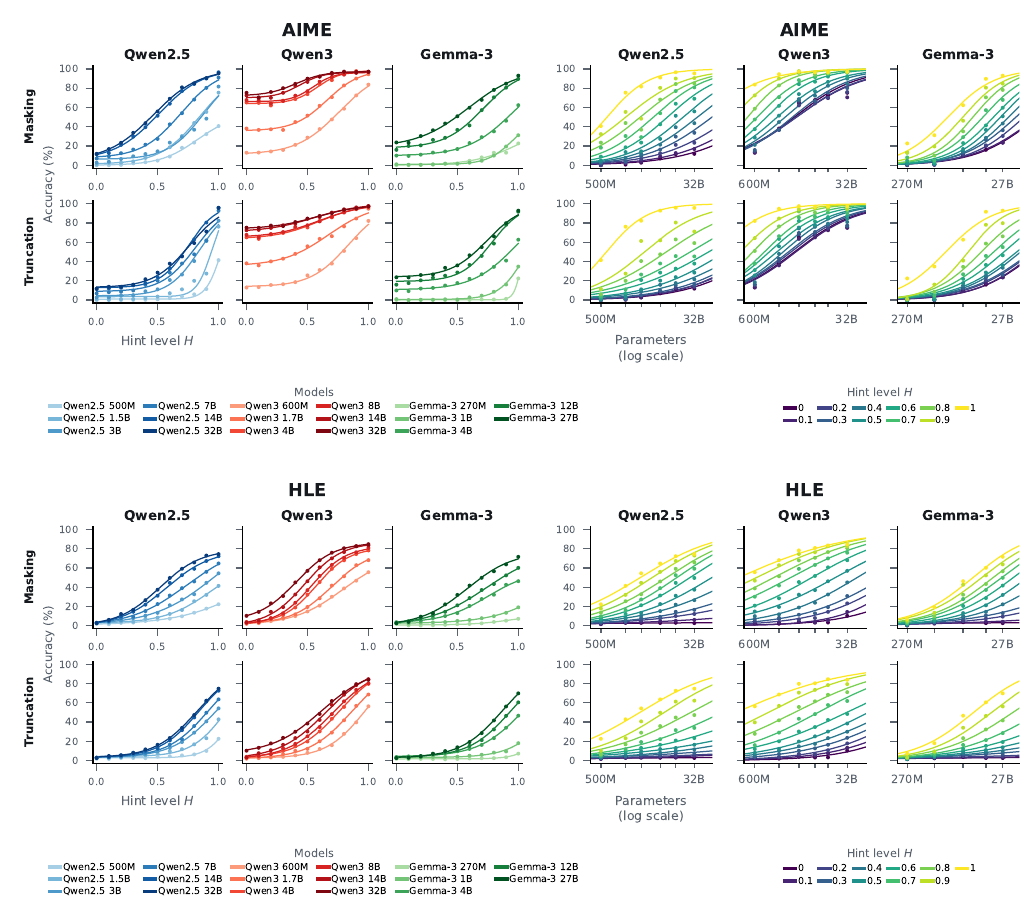}
    \caption{\textbf{Hint and capability slices for the remaining
    families.} As \cref{fig:phenomena_ladders}, for Qwen2.5, Qwen3 and
    Gemma-3 on AIME (top) and HLE (bottom). Rows are masking and truncation.
    \textbf{Left:} accuracy against hint level $H$, one curve per model,
    darker is larger. \textbf{Right:} accuracy against model scale, one
    curve per hint level. Each curve is a sigmoid fitted to that model or
    that level alone, with free asymptotes in $H$ and asymptotes pinned to
    0 and 1 in scale.}
    \label{fig:all_families_ladders}
\end{figure}

\begin{figure}[htbp]
    \centering
    \includegraphics[width=0.8\linewidth]{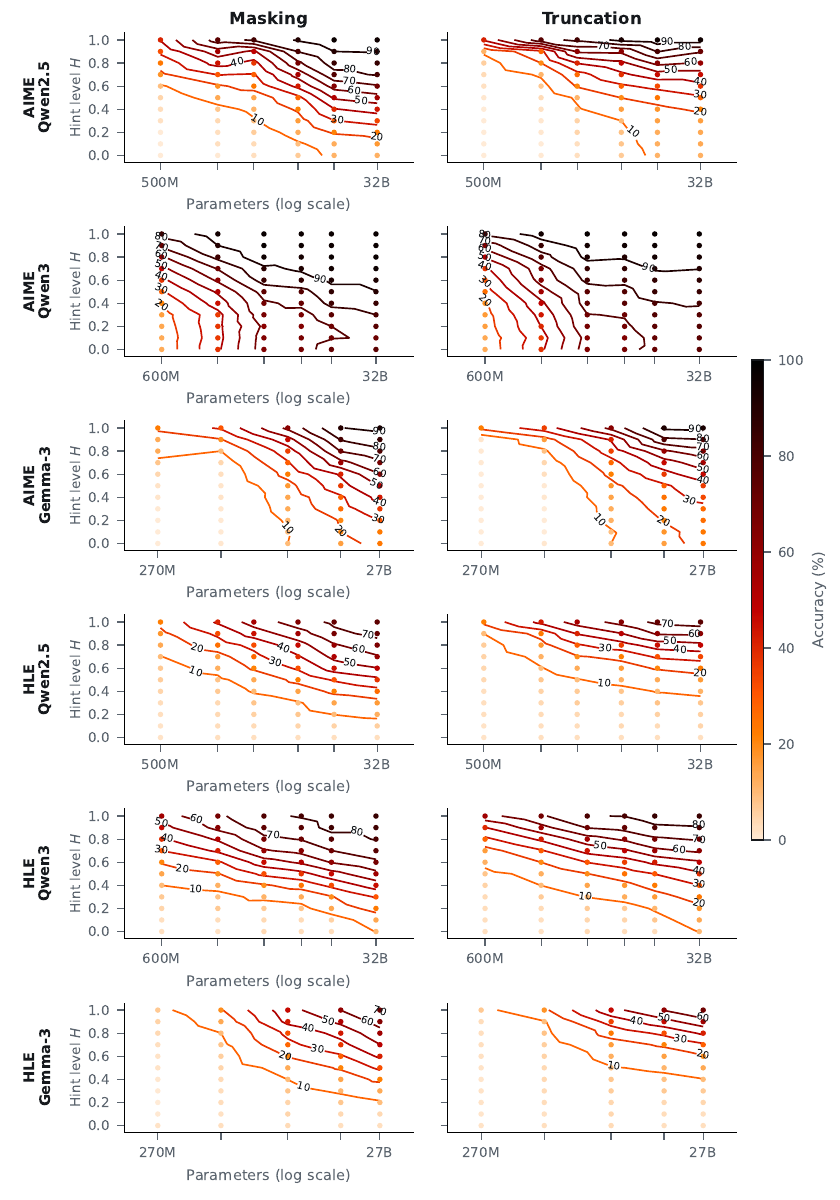}
    \caption{\textbf{Iso-accuracy contours for the remaining families.} As
    \cref{fig:iso_accuracy}, for Qwen2.5, Qwen3 and Gemma-3 on AIME and
    HLE. Columns are masking and truncation.}
    \label{fig:all_families_contours}
\end{figure}

\cref{sec:phenomena} shows the Delphi ladder and Qwen3.5 on HLE. Here we
show the same analyses for the remaining (benchmark, family) pairs:
Qwen2.5, Qwen3 and Gemma-3 on both AIME and HLE.
\cref{fig:all_families_ladders} gives the hint and capability slices,
with each slice fitted alone as in \cref{fig:phenomena_ladders}, and
\cref{fig:all_families_contours} gives the empirical iso-accuracy contours,
as in \cref{fig:iso_accuracy}. The same patterns hold. Accuracy rises
smoothly with the hint level for every model, the ordering of each family
is preserved, and the contours are smooth.

\section{Additional Forecasting Results}
\label{app:benchmark_performance}

\begin{figure}[htbp]
    \centering
    \includegraphics[width=\linewidth]{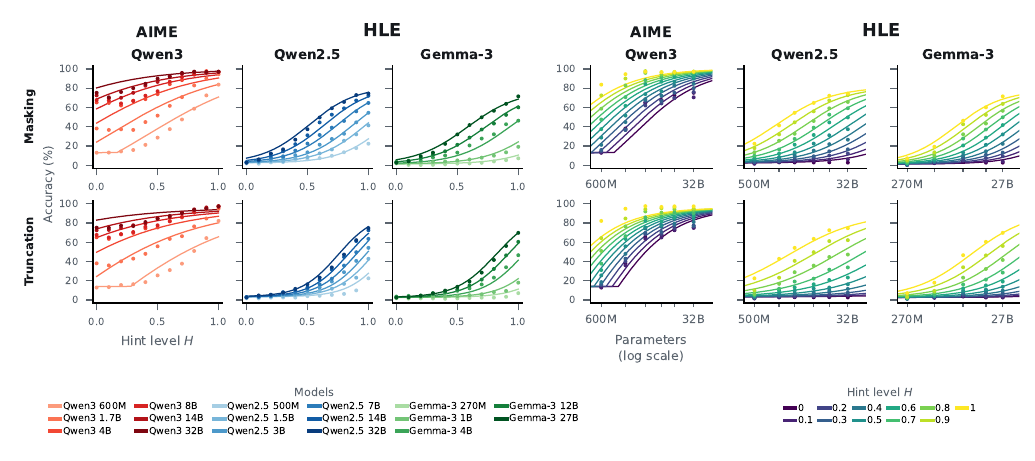}
    \caption{As \cref{fig:posttraining-well-spec}, for Qwen3 on AIME and Qwen2.5 and
    Gemma-3 on HLE.}
    \label{fig:missing_joint}
\end{figure}

\begin{figure}[htbp]
    \centering
    \includegraphics[width=\linewidth]{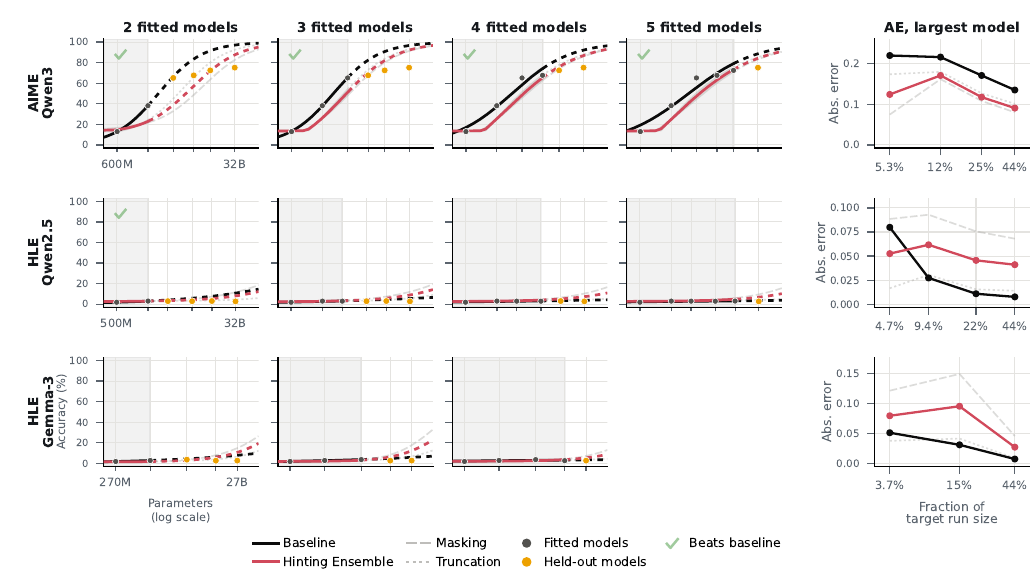}
    \caption{As \cref{fig:postraining-identifiable}, for Qwen3 on AIME and Qwen2.5 and
    Gemma-3 on HLE.}
    \label{fig:missing_forecasts}
\end{figure}

\cref{fig:missing_joint,fig:missing_forecasts} show the joint-law fits and
the forecasts for the (benchmark, family) pairs not reported in
\cref{sec:benchmark_performance}: Qwen3 on AIME, and Qwen2.5 and Gemma-3
on HLE. We did not run Qwen3.5 on AIME.

\begin{figure}[htbp]
    \centering
    \includegraphics[width=\linewidth]{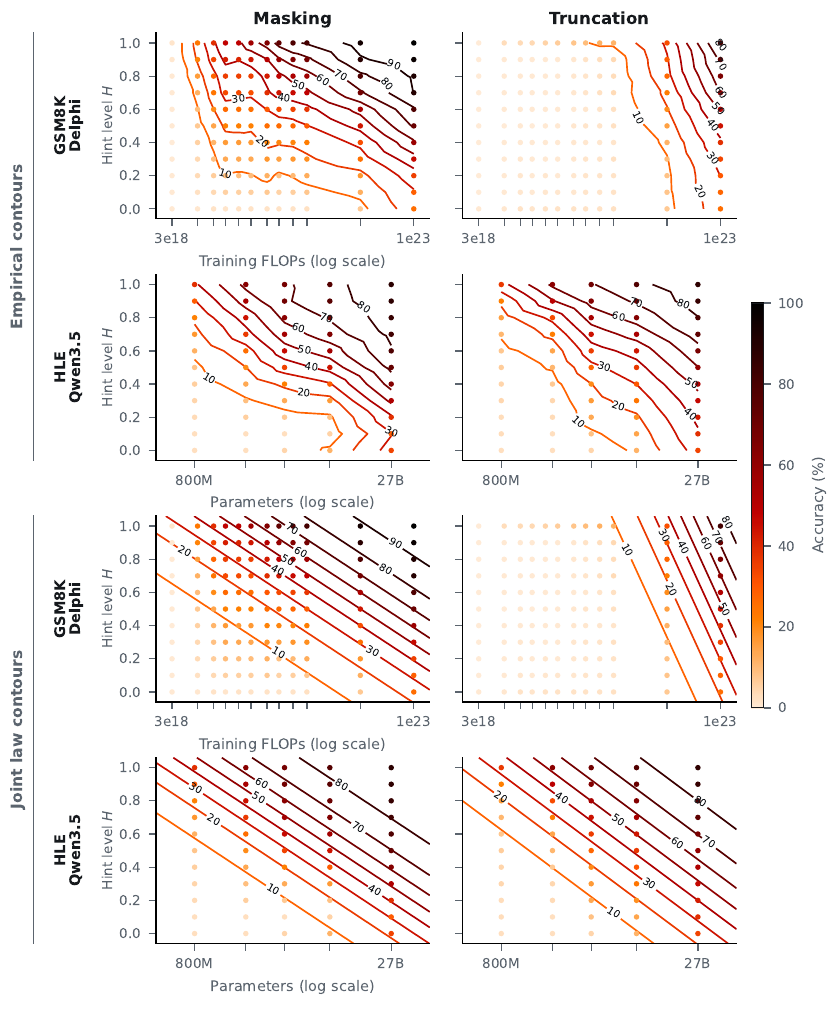}
    \caption{\textbf{Empirical against joint-law iso-accuracy contours.}
    \textbf{Top two rows:} \cref{fig:iso_accuracy}, the Delphi ladder on 
    GSM8K and the Qwen3.5 family on HLE with contours interpolated from 
    the measured cells. \textbf{Bottom two rows:} the same cells with the 
    contours of the joint law of \cref{eq:joint_law} fitted to each 
    panel, which above the law's floor are straight parallel lines of slope $-\beta_1 / \beta_2$. 
    Columns are masking and truncation. Each dot is one measured 
    cell colored by accuracy.}
    \label{fig:contours_empirical_vs_joint}
\end{figure}

\cref{fig:contours_empirical_vs_joint} compares the empirical 
iso-accuracy contours of \cref{fig:iso_accuracy} with the contours 
implied by the joint law of \cref{eq:joint_law}. Under that law every 
contour is a straight line with the same slope $-\beta_1 / \beta_2$, 
so the lower half of the figure shows what the surface would look like 
if the law held exactly above its floor. For Qwen3.5 on HLE the empirical 
contours are close to the implied ones for both masking and truncation. 
For the Delphi ladder on GSM8K the implied contours also track the
empirical ones for both methods, with the largest deviations near
the floor, where the empirical contours of masking
flatten across the smaller models. These deviations are the reason we describe the joint law
as an approximation.

\subsection{Two-Stage Scaling Law Baseline}
\label{app:two_stage}

A natural alternative to \method{} is to forecast through an intermediate
\emph{soft} metric rather than through a hinted benchmark. The soft
metric is a continuous quantity that keeps moving while accuracy is still
pinned at the floor, and the forecast is a chain of two fits, compute to
the soft metric, then the soft metric to accuracy. \citet{held2026delphi}
uses exactly this procedure on the Delphi ladder, combining a compute
scaling law with an \emph{observational} scaling law 
\citep{ruan2024observationalscalinglawspredictability}. We test this method here as a baseline.

Following \citet{held2026delphi}, we use bits-per-byte on the GSM8K solution 
trace as the soft metric averaged over problems. We use the 
same standard 5-shot prompt.
We restrict this comparison to the ten compute budgets and single
seed that \citet{held2026delphi} uses. All accuracies in this section, including
those of the \method{} forecasts we compare against, are measured on
these same ten models.
As a check on our implementation, our bits-per-byte measurements track those reported by
\citet{held2026delphi} for the same ladder closely, differing by $0.84\%$ on
average and $2.09\%$ at worst across all ten rungs, despite differences in
problem subset and prompt details.

We then fit two scaling laws and compose them. The first maps 
model scale $C$ to bits-per-byte with a power law in FLOPs, and the second 
maps bits-per-byte to accuracy with a sigmoid. Concretely:
\begin{align}
\mathrm{BPB}(C) &= L_\infty + A\,10^{-\alpha C},
\label{eq:bpb_power_law}\\
\mathrm{Acc}(M, 0) &= \sigma\!\left(b_1 \mathrm{BPB}(C_M) + b_0\right),
\label{eq:bpb_sigmoid}
\end{align}
where $C_M$ is the scale of model $M$.

\begin{figure}[htbp]
    \centering
    \includegraphics[width=0.9\linewidth]{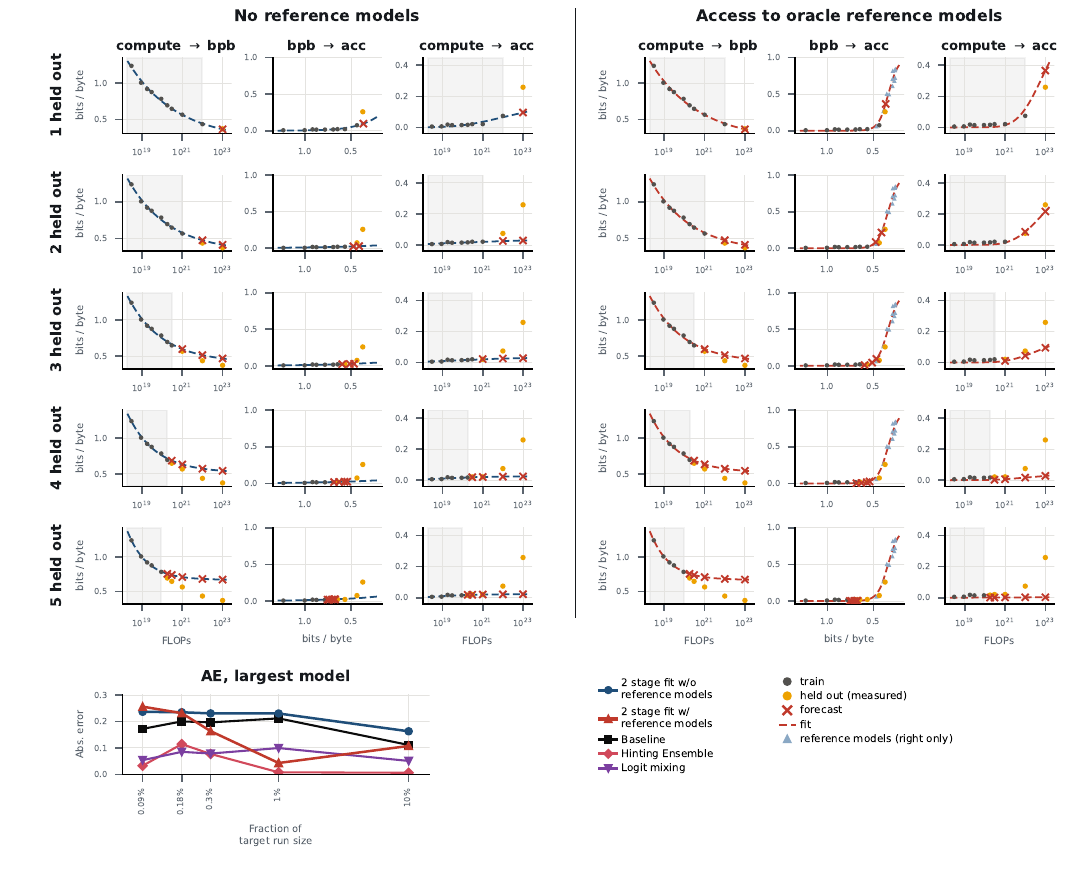}
    \caption{\textbf{Two-stage scaling law forecasts on the Delphi ladder
    (GSM8K).} Each row is fitted on the smallest models only (the
    \textbf{shaded} region) and forecasts the held-out larger ones.
    \textbf{Columns within a block:} the compute-to-$\mathrm{BPB}$ power law
    of \cref{eq:bpb_power_law}; the $\mathrm{BPB}$-to-accuracy sigmoid of
    \cref{eq:bpb_sigmoid}, with the $x$ axis running high to low so the
    curve rises left to right; and the two composed, giving accuracy implied
    at any compute. \traindot\ are fitted models and \testdot\ held-out
    models; \textbf{\textcolor{fcstcol}{crosses}} are forecasts, which use
    only the held-out model's compute. \textbf{Left block:} the sigmoid is
    fitted on the fitted models alone. \textbf{Right block:} the eleven
    external reference models of \citet{held2026delphi}
    (\textbf{\textcolor{refcol}{triangles}}) are added to its fitting
    data. \textbf{Bottom:} absolute error for the largest model on the 0--1
    accuracy scale against the size of the largest fitted model as a
    fraction of the target's, with
    the unhinted baseline, the hinting ensemble of
    \cref{fig:gsm8k-delphi-downscaling} and logit mixing
    (\cref{sec:logit-mixing}) for reference, all refitted on the same ten
    models.}
    \label{fig:two_stage_scaling_law}
\end{figure}

We consider two different ways of fitting these laws and 
creating a forecast. The most comparable to our 
results in \cref{fig:gsm8k-delphi-downscaling} is 
to fit both on the $n$ smallest models
in the ladder, and evaluate on the held-out 
larger models. The results of doing this are shown on the 
left of \cref{fig:two_stage_scaling_law}, with 
the absolute error of the prediction shown in blue 
on the bottom plot. We find this method 
does \emph{very poorly}, predicting 
essentially a flat line at 0, resulting in a roughly 
constant absolute error of 16--24 percentage points. This makes 
perfect sense because the second stage 
of the fit in \cref{eq:bpb_sigmoid} is 
simply fitting a sigmoid to a row of near zero 
accuracy values, the exact same problem 
as the $H=0$ sigmoid we use as a baseline in the main paper.

We also consider a second setting used by \citet{held2026delphi}.
In this case we imagine we have access to a set of 
stronger \emph{reference} pretrained models, which already 
exceed the accuracy of the largest model we plan on scaling to. It is important 
to note that this assumption is \emph{not realistic} if 
one is running a frontier training run beyond the scale 
of any publicly available base model. However, 
if we assume this is possible, we can measure 
these models' benchmark accuracies and bits-per-byte,
and use them to anchor the 
bits-per-byte to accuracy sigmoid. 

We show results for this reference-model 
fit on the right of \cref{fig:two_stage_scaling_law}.
We find that the sigmoid is indeed fitted well; 
however, with enough held-out models the 
compute to bits-per-byte scaling law ends up 
being inaccurate, leading also to poor predictions 
overall. It is worth emphasizing that 
the downstream accuracy prediction is \emph{very}
sensitive to even small errors in the bits-per-byte 
prediction, as the three held-out models in \cref{fig:two_stage_scaling_law} show.
Indeed, the model we wish to predict, at $10^{23}$ FLOPs, has only just emerged 
on the bits-per-byte to accuracy sigmoid, and so it makes sense 
that small bits-per-byte errors would cause large accuracy errors 
as the true $10^{23}$ model sits close to the steepest point 
of the sigmoid.

\begin{figure}[htbp]
    \centering
    \includegraphics[width=\linewidth]{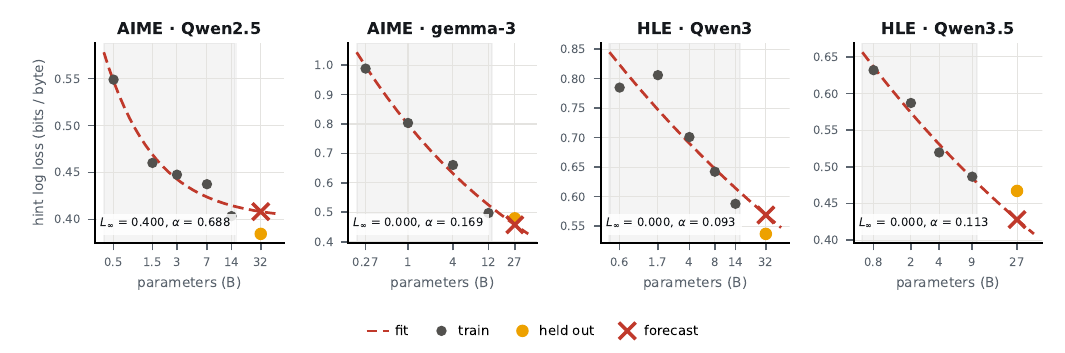}
    \caption{\textbf{Bits-per-byte against parameter count on the
    post-trained families.} Each panel holds out the largest model and fits
    \cref{eq:bpb_power_law} to the rest (the \textbf{shaded} region), then
    reads the fit at the held-out model's parameter count. \traindot\ are
    fitted models, \testdot\ the held-out model and
    \textbf{\textcolor{fcstcol}{crosses}} the forecast. Fitted $L_\infty$
    and $\alpha$ are annotated per panel; $L_\infty$ reaches its lower bound
    of zero in three of the four families, indicating the floor is not
    identified at this number of points.}
    \label{fig:posttrained_power_law}
\end{figure}

Finally, as expected, bits-per-byte scales much less cleanly for the post-trained
families than for the Delphi ladder, as shown in \cref{fig:posttrained_power_law},
making the method infeasible for these ladders.

\subsection{Pass@$k$ as a Hinting Method}
\label{app:pass_at_k}

Here we consider pass@$k$ as a hinting method. 
\emph{Pass@$k$} marks a problem as solved if any of $k$
independently sampled attempts is correct, so $k = 1$ is the ordinary
evaluation and accuracy rises monotonically in $k$. 

It is easy to see that pass@$k$ is a valid \method{} method.
It is
indexed by a scalar, it returns the unmodified benchmark at one end of its
range, and it raises accuracy monotonically as it is turned up. We
therefore treat it as one more \method{} method, with hint level
$H = \log_2 k$
so that $H = 0$ is the unmodified benchmark and each additional unit of $H$
doubles the sampling budget. We then fit the same joint law of
\cref{eq:joint_law} over the resulting grid of (scale, hint) pairs 
and forecast exactly as before, by evaluating the fitted law's $H = 0$
slice at the held-out scales. 

\begin{figure}[htbp]
    \centering
    \includegraphics[width=0.85\linewidth]{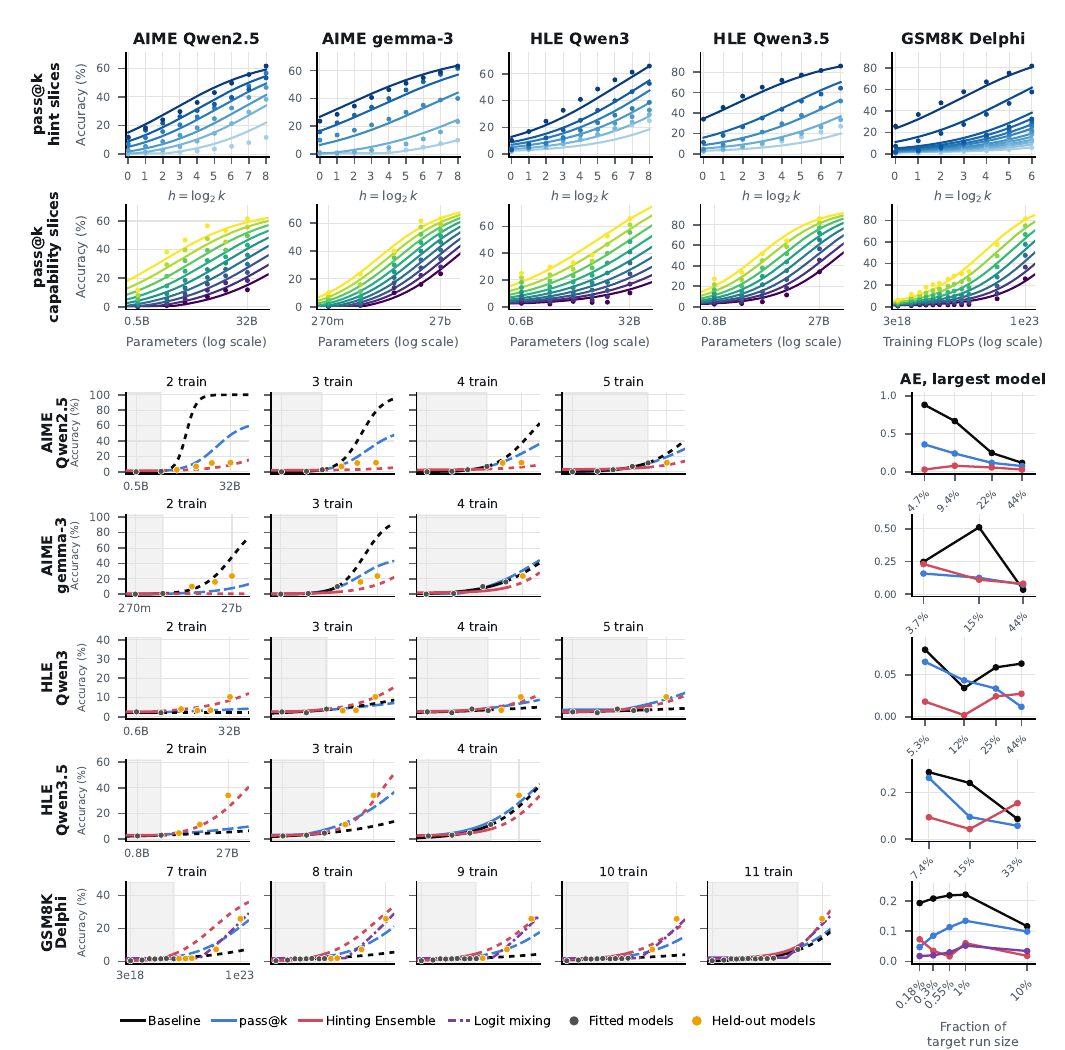}
    \caption{\textbf{Pass@$k$ as a \method{} method.} \textbf{Top two
    rows:} the joint law of \cref{eq:joint_law} fitted to the pass@$k$ grid
    of each (benchmark, family) pair, drawn as its hint slices
    (accuracy against $H = \log_2 k$, one line per model, darker with
    scale) and its capability slices (accuracy against scale, one line
    per hint level, brighter with $H$). \textbf{Bottom:} the same law
    refitted on the $n$ smallest models only (the shaded region)
    and evaluated at $H = 0$ on the held-out larger ones. 
    We compare against the hinting ensemble of masking and truncation
    (\cref{sec:mask-trunc}), logit mixing on the Delphi ladder
    (\cref{sec:logit-mixing}), and the regular 
sigmoid fitted to the $H=0$ accuracies.}
    \label{fig:pass_at_k}
\end{figure}

\cref{fig:pass_at_k} shows the result. The joint law describes the pass@$k$
slices well. In terms of forecasting, the method 
generally improves on the baseline. However, it underperforms our main 
ensembling method.

\section{When Can Hinting Inform a Forecast?}
\label{app:unassisted_determined}

\begin{figure}[htbp]
    \centering
    \includegraphics[width=0.85\linewidth]{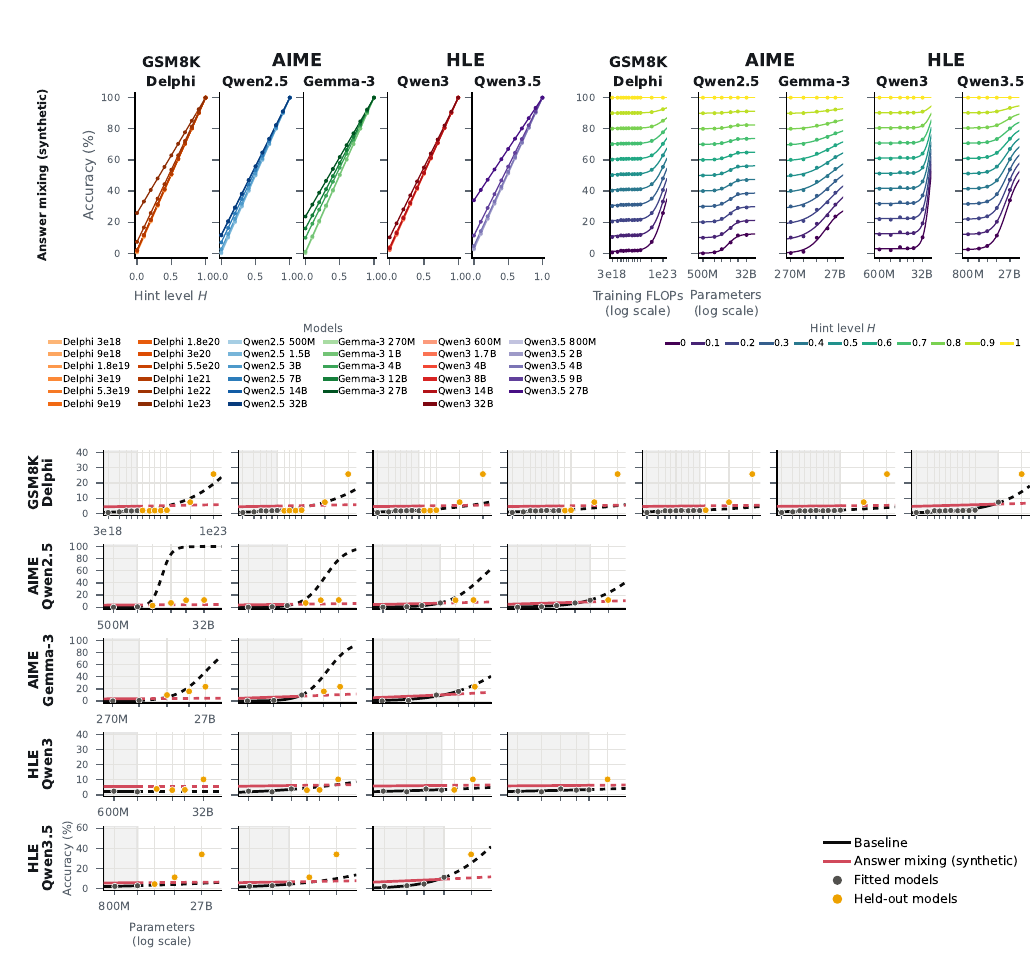}
    \caption{\textbf{Answer mixing carries no information for forecasting.}
    \textbf{Top:} hint and capability slices of answer mixing, generated
    from each family's unhinted accuracies as
    $\mathrm{Acc}(M, H) = H + (1-H)\,\mathrm{Acc}(M, 0)$. Each hint slice is
    fitted with a straight line and each capability slice with a sigmoid
    with free asymptotes. \textbf{Bottom:} forecasts of the largest model's
    unhinted accuracy from the joint law of \cref{eq:joint_law} fitted to
    the answer-mixing cells of the smallest models (red), against the
    baseline sigmoid fitted to the unhinted scores alone (black). The
    shaded region spans the fitted models, grey dots are fitted models and
    orange dots held-out models.}
    \label{fig:answer_mixing}
\end{figure}

Not every way of making a benchmark easier helps with forecasting. As a
control, consider \emph{answer mixing}, a hinted protocol $\omega_H$ that
returns the reference answer with probability $H$ and otherwise samples
from the model as usual. Its accuracy is
$\mathrm{Acc}(M, H) = H + (1-H)\,\mathrm{Acc}(M, 0)$, so every hint slice
is a straight line from the model's unhinted score to 1.
In \cref{fig:answer_mixing} we show empirically how this 
method performs.

From \cref{fig:answer_mixing} we see that the method 
performs terribly for forecasting. Intuitively 
this makes sense. Because the hint slices 
only depend on the accuracy of the model 
at $H=0$, the hinted curves of all 
models with $\mathrm{Acc}(M,0) = 0$ perfectly overlap.
This means that every set of pre-emergence 
models \emph{produces the same forecast}, which 
clearly cannot be correct.

From this, we get better intuition as to what makes 
a good \method{} method, \emph{one that 
separates pre-emergence models according
to the difference in their underlying capability}.
Below we formalize this property of overlapping $H=0$ 
curves, and how such \method{} methods cannot exceed 
the performance of simply predicting based on 
unhinted benchmark scores.

\paragraph{Setup.}
A ladder has models $M_1, \dots, M_n$ at scales $C_1 < \dots < C_n$ and a
target $M_\star$ at $C_\star > C_n$. Write $a_i(H) = \mathrm{Acc}(M_i, H)$ and
$a_\star = \mathrm{Acc}(M_\star, 0)$. A forecaster observes a fixed, known grid of
hint levels $\mathcal{H}$ containing $0$ and the data
\begin{equation*}
D = \{(C_i, H, a_i(H)) : i \le n,\ H \in \mathcal{H}\},
\end{equation*}
and outputs $\hat a_\star = f(D)$. The standard baseline sees only the
unhinted data $D_0 = \{(C_i, a_i(0)) : i \le n\}$. We call a \method{}
method \emph{unhinted-determined} if there is a fixed function $g$, the same
for every model, such that
\begin{equation}
\label{eq:unassisted_determined}
\mathrm{Acc}(M, H) = g\big(H, \mathrm{Acc}(M, 0)\big) \quad \text{for all models } M \text{ and levels } H.
\end{equation}
Answer mixing is unhinted-determined with $g(H, a) = H + (1-H)a$.

\paragraph{Proposition 1.}
\emph{Let the ladder, and with it $D$, $D_0$ and $a_\star$, be drawn from any
distribution $P$. For an unhinted-determined method, every forecaster $f$ of $D$
has a forecaster $f_0$ of $D_0$ with $f_0(D_0) = f(D)$ on every ladder.
Consequently, for any loss $\ell$,}
\begin{equation*}
\inf_f \mathbb{E}_P\,\ell\big(f(D), a_\star\big) = \inf_{f_0} \mathbb{E}_P\,\ell\big(f_0(D_0), a_\star\big),
\end{equation*}
\emph{and the same holds with the expectation replaced by the worst case over
ladders.}

\emph{Proof.} By \cref{eq:unassisted_determined}, $D = \Phi(D_0)$, where $\Phi$
replaces each $(C_i, a_i(0))$ by $\{(C_i, H, g(H, a_i(0)))\}_{H \in \mathcal{H}}$
and depends only on the known $g$ and $\mathcal{H}$. Take $f_0 = f \circ \Phi$.
Since $f$ and $f_0$ give the same output on every ladder, they have the
same loss on every ladder, hence the same expected loss under $P$ and the
same worst-case loss. So the best forecaster of $D$ is no better than the
best forecaster of $D_0$. Conversely $D$ contains the unhinted cells
(as $0 \in \mathcal{H}$), so it is no worse either. $\qquad\square$

In particular, fitting the joint law of \cref{eq:joint_law} to the
hinted cells of an unhinted-determined method is just one particular
forecaster of $D_0$.
When every fitted model scores zero without hints, the
unhinted data leave the target's accuracy completely unconstrained, so
by Proposition 1 the hinted data of an unhinted-determined method
leave it unconstrained too.

\section{Model Ranking Details}
\label{app:model_ranking}

\subsection{Model Sets and Ground Truth}
\label{app:ranking_ground_truth}
To be able to compare ranking methods, we need to have a ground truth to compare to. Here, we describe how we determine that ground truth and how we select the set of models to rank. 

We use AIME and HLE as the benchmarks. On each benchmark we start from a pool of post-trained models from
several families, 32 on AIME and 25 on HLE. 

\paragraph{Collecting rollouts.} 
To get an accurate ground-truth ranking, we compute an avg@$k$ score with very large $k$. 
Every model has two disjoint pools of unhinted rollouts per problem: a method pool of 128 on AIME and 64 on HLE, which is the only unhinted data the ranking methods draw from, and a ground-truth pool used solely to compute the ground-truth ranking. 
The ground-truth pool starts at 128 rollouts per problem on AIME and 192 on HLE. We then deepen it for models that score near zero: those below $1\%$ on AIME, 14 models, receive 4{,}096
ground-truth rollouts per problem, and those below $3\%$ on HLE, 16
models, receive 1{,}024, so counting the method pool these models have 4{,}224 and 1{,}088 unhinted rollouts per problem in total. We use fewer rollouts on HLE because its
grading requires a paid LLM judge. We only collect this deeper sample on part of the original model sets because it is compute-heavy and expensive to run. Deepening moves the estimates
slightly, and one HLE model that was below $3\%$ at 192 samples
ends at $3.0\%$.

\paragraph{Creating a stable ranking and choosing a subset of models.} Simply ranking models by avg@$k$ with high $k$ does not guarantee a ranking is correct. Thus, we choose only a subset of the models, such that every pair of models in the set is certified at the 95\% level with a Bonferroni correction over
all pairs in the pool. 
We treat the
benchmark problems as fixed, so the uncertainty is rollout sampling
noise only, and form a confidence interval for each pair's accuracy difference
from binomial standard errors computed from the observed per-problem
accuracies. 
A pair is certified if the Bonferroni-corrected 95\% confidence interval on the accuracy difference between two models does not include zero. This is desirable because if the confidence interval on the accuracy difference between two models does not include zero, their ranking cannot flip.
Because the intervals are simultaneous over all pairs, the Bonferroni-corrected guarantee
holds for any subset of models chosen afterwards, so no further
correction is needed for the choice of a model set from this model pool.

From the certified model pairs we take the
longest chain of models in which every consecutive pair is certified and keep its seven
lowest-accuracy members. The chosen
sets are marked in \cref{tab:ranking_models}. 
These model sets are both near zero and nearly indistinguishable: their unhinted accuracies are
at most $0.81\%$ on AIME and $3.0\%$ on HLE, making them near zero. They span $0.06\%$ to $0.81\%$ on AIME and $1.7\%$ to $3.0\%$ on HLE, less than 1 and 2 percentage points respectively, making them nearly indistinguishable.

\begin{table}[htbp]
\caption{Model pools for the ranking experiments. Checkmarks mark the evaluated set. Rollouts counts the ground-truth unhinted rollouts per problem.}
\label{tab:ranking_models}
\begin{center}
\footnotesize
\setlength{\tabcolsep}{4pt}
\begin{tabular}[t]{lrrc}
\multicolumn{4}{c}{\bf AIME} \\
\toprule
\multicolumn{1}{l}{\bf Model} & \multicolumn{1}{c}{\bf Acc.\ (\%)} & \multicolumn{1}{c}{\bf Rollouts} & \multicolumn{1}{c}{\bf Set} \\
\midrule
Qwen3 32B & 75.091 & 128 &  \\
Gemma-4 31B & 73.984 & 128 &  \\
Qwen3 14B & 72.201 & 128 &  \\
Qwen3 30B-A3B & 71.927 & 128 &  \\
Gemma-4 26B-A4B & 71.003 & 128 &  \\
Qwen3 8B & 68.086 & 128 &  \\
Qwen3 4B & 64.531 & 128 &  \\
Qwen3 1.7B & 37.266 & 128 &  \\
Gemma-4 E4B & 36.901 & 128 &  \\
Gemma-4 E2B & 28.789 & 128 &  \\
Gemma-3 27B & 22.734 & 128 &  \\
Gemma-3 12B & 17.161 & 128 &  \\
Qwen2.5 32B & 13.281 & 128 &  \\
Qwen3 600M & 12.917 & 128 &  \\
Qwen2.5 14B & 11.471 & 128 &  \\
Gemma-3 4B & 10.534 & 128 &  \\
Qwen2.5 7B & 7.669 & 128 &  \\
Qwen2.5 3B & 3.255 & 128 &  \\
Llama-3.1 8B & 0.812 & 4,096 & \checkmark \\
Qwen2.5 1.5B & 0.665 & 4,096 & \checkmark \\
Gemma-3 1B & 0.484 & 4,096 & \checkmark \\
Gemma-1.1 7B & 0.247 & 4,096 & \checkmark \\
Llama-3.2 1B & 0.222 & 4,096 &  \\
Gemma-2 9B & 0.199 & 4,096 &  \\
Qwen2.5 500M & 0.175 & 4,096 & \checkmark \\
Mistral 7B & 0.110 & 4,096 & \checkmark \\
Qwen1.5 1.8B & 0.107 & 4,096 &  \\
Gemma-3 270M & 0.103 & 4,096 &  \\
Gemma-1.1 2B & 0.101 & 4,096 &  \\
Llama-2 7B & 0.089 & 4,096 &  \\
Gemma-2 2B & 0.076 & 4,096 &  \\
Llama-2 13B & 0.059 & 4,096 & \checkmark \\
\bottomrule
\end{tabular}\qquad
\begin{tabular}[t]{lrrc}
\multicolumn{4}{c}{\bf HLE} \\
\toprule
\multicolumn{1}{l}{\bf Model} & \multicolumn{1}{c}{\bf Acc.\ (\%)} & \multicolumn{1}{c}{\bf Rollouts} & \multicolumn{1}{c}{\bf Set} \\
\midrule
Gemma-4 31B & 18.700 & 192 &  \\
Gemma-4 26B-A4B & 17.324 & 192 &  \\
Qwen3 32B & 10.612 & 192 &  \\
Qwen3 30B-A3B & 6.274 & 192 &  \\
Qwen3 14B & 4.273 & 192 &  \\
Gemma-4 E4B & 3.962 & 192 &  \\
Gemma-3 4B & 3.616 & 192 &  \\
Qwen3 4B & 3.527 & 192 &  \\
Qwen2.5 3B & 3.220 & 192 &  \\
Gemma-4 E2B & 3.006 & 1,024 & \checkmark \\
Qwen2.5 14B & 2.957 & 1,024 &  \\
Qwen2.5 7B & 2.898 & 1,024 &  \\
Gemma-3 27B & 2.828 & 1,024 & \checkmark \\
Qwen3 8B & 2.772 & 1,024 &  \\
Llama-2 13B & 2.764 & 1,024 &  \\
Qwen2.5 1.5B & 2.624 & 1,024 & \checkmark \\
Gemma-3 12B & 2.619 & 1,024 &  \\
Qwen3 600M & 2.575 & 1,024 &  \\
Gemma-3 1B & 2.521 & 1,024 &  \\
Qwen2.5 32B & 2.431 & 1,024 & \checkmark \\
Llama-3.1 8B & 2.241 & 1,024 & \checkmark \\
Qwen3 1.7B & 2.220 & 1,024 &  \\
Qwen2.5 500M & 2.124 & 1,024 &  \\
Gemma-3 270M & 2.001 & 1,024 & \checkmark \\
Llama-2 7B & 1.658 & 1,024 & \checkmark \\
\bottomrule
\end{tabular}
\end{center}
\end{table}

\subsection{The Hint-Prior Ranking}
\label{app:ranking_hint_prior}

Given a budget of $k$ rollouts per problem, the hint-prior ranking
spends one rollout per problem on hinted benchmarks and the
remaining $k - 1$ at $H = 0$. Each problem's hinted rollout uses a
single hint level in $\{0.1, \ldots, 1.0\}$, dealt in turn over a random
problem order, so the levels get equal shares up to rounding. The hinted rollouts average to the hinted
score $\widehat{\mathrm{Acc}}_1(M, \bar H)$, where $\bar H$ is
the pool of hint levels; the unhinted rollouts average to the
unhinted score $\widehat{\mathrm{Acc}}_{k-1}(M, 0)$, as in avg@$k$.
Both scores are standardized across models before being combined
in \cref{eq:assistance_prior}. If all models tie on a score, its z-scores are zero and the other decides. We use Kendall's $\tau$-a.

The weighting reflects the two scores' errors. The unhinted score is
unbiased but noisy, and its noise shrinks with $k$. The hinted score
is far less noisy but biased if hinted rankings do not match
the ground-truth unhinted ranking. This bias does not shrink with $k$. Therefore, the weight
$\alpha_k$ moves from the hinted score to the unhinted one
as the budget grows. Because $\alpha_k$ depends only on the budget
and not on the data, the hint-prior ranking becomes avg@$k$ at large
budgets even when the hinted ranking is wrong, which happens for some
hinting methods, as \cref{app:ranking_other_methods} shows. Below one rollout
per problem there is no unhinted score, and the single hinted rollout
covers a random subset of the problems.

\paragraph{Choice of $\lambda$.}\label{app:assistance_prior}

\begin{figure}[htbp]
    \centering
    \includegraphics[width=0.8\linewidth]{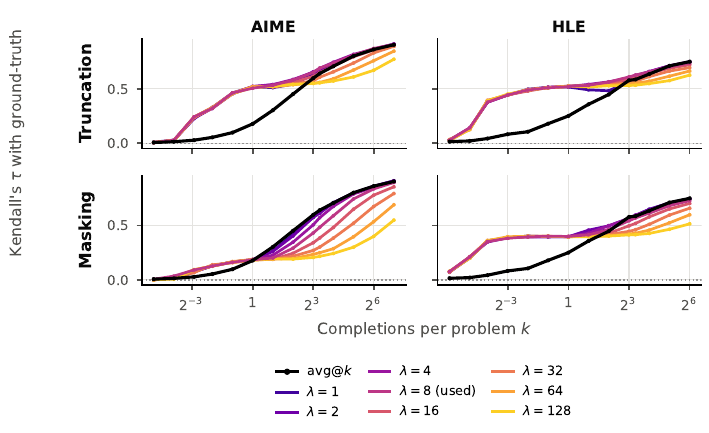}
    \vspace{-6mm}
    \caption{\textbf{Sensitivity of the hint-prior ranking to $\lambda$.} Kendall's $\tau$ against the budget $k$ for the hint-prior ranking (\cref{eq:assistance_prior}) with $\lambda \in \{1, 2, 4, 8, 16, 32, 64, 128\}$, for both hinting methods (rows) on both benchmarks (columns), with avg@$k$ in black. $\lambda$ is the number of unhinted rollouts per problem the hinted score is worth: small $\lambda$ hands over to the unhinted score early, large $\lambda$ late.}
    \label{fig:lambda_sweep}
\end{figure}

The weight $\alpha_k = (k-1) / (k-1+\lambda)$ in \cref{eq:assistance_prior} treats the hinted score as a prior worth $\lambda$ unhinted rollouts per problem. \cref{fig:lambda_sweep} sweeps $\lambda$ from $1$ to $128$. No single $\lambda$ is optimal. Small $\lambda$ values shift weight to the unhinted score quickly, which is suboptimal if the hinted scores are correct, but large $\lambda$ values keep weight on the hinted scores for too long, which is suboptimal if the hinted scores are incorrect. 
We use $\lambda = 8$ throughout as an intermediate value.

\subsection{Ranking with Other Hinting Methods}
\label{app:ranking_other_methods}

\begin{figure}[htbp]
    \centering
    \includegraphics[width=1\linewidth]{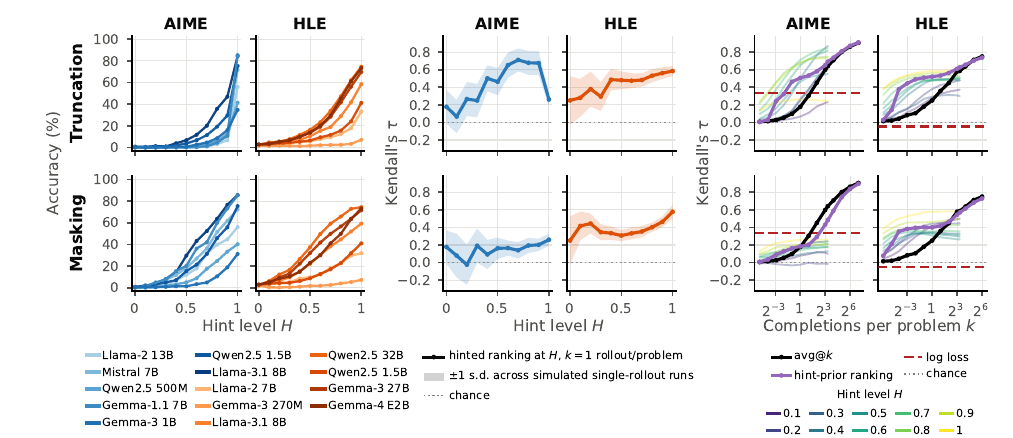}
    \vspace{-6mm}
    \caption{\textbf{\cref{fig:model_ranking} for truncation and masking.} Rows are hinting methods, columns benchmarks; left, middle and right panels as in \cref{fig:model_ranking}.}
    \label{fig:model_ranking_all_methods}
\end{figure}

\cref{fig:model_ranking_all_methods} repeats \cref{fig:model_ranking} for both truncation and masking.
Hinted rankings largely agree with the ground-truth unhinted ranking under every method: in three of the four method--benchmark pairs, a single hinted rollout per problem ranks the models closer to the ground truth than a single unhinted one, and often by a wide margin (middle panels). Truncation is the more reliable method, with a median $\tau$ over hint levels of $0.48$ on AIME and $0.48$ on HLE. Masking reaches $0.38$ on HLE but is no better than an unhinted rollout on AIME, where families gain from the hint at different rates and the curves cross.

The hint-prior ranking follows this agreement (right panels). On HLE, where the hinted ranking of masking is informative, it leads avg@$k$ at small budgets, by $0.15$ at one rollout per problem. On AIME, where it is not, it trails avg@$k$ at intermediate budgets before converging with it. Wherever hinted rankings agree with unhinted ones, the hint-prior ranking recovers the ranking from a fraction of the rollouts avg@$k$ needs, and where they do not, it falls back to avg@$k$ as the budget grows.

\end{document}